\documentclass[10pt,twocolumn,letterpaper]{article}

\usepackage[pagenumbers]{cvpr} 

\usepackage{enumitem}
\usepackage{booktabs}
\usepackage{multirow}
\usepackage{color, colortbl}
\usepackage{subcaption}
\usepackage{pifont}
\usepackage{makecell}
\usepackage{threeparttable}
\usepackage{hhline}
\usepackage{scalerel,xparse}
\usepackage{rotating} 

\usepackage{tablefootnote}
\usepackage{minted}
\usepackage{pdflscape}

\usepackage{tabularx}
\usepackage{lipsum}
\usepackage{listings}
\usepackage{amsmath}
\usepackage{amssymb}

\newfloat{Algorithm}{htbp}{loa}

\definecolor{LightYellow}{rgb}{1.0, 1.0, 0.88} 
\definecolor{LightBlue}{rgb}{0.88, 1.0, 1.0}
\definecolor{LightGreen}{rgb}{0.9, 0.9, 1.0}
\definecolor{Gray}{gray}{0.5}
\definecolor{LGray}{gray}{0.9}
\definecolor{darkblue}{RGB}{94,110,186}
\definecolor{darkGreen}{RGB}{92, 148, 110}
\definecolor{myblue}{RGB}{14, 121, 178}
\definecolor{myred}{RGB}{192, 0, 0}

\newcommand{\jy}[1]{\textcolor{violet}{#1}}

\newcommand{\ysqq}[1]{\textcolor{darkGreen}{#1}}

\renewcommand{\jy}[1]{#1}

\renewcommand{\ysqq}[1]{#1}

\newcommand{\cmark}{\ding{51}}%
\newcommand{\xmark}{\ding{55}}%

\definecolor{cvprblue}{rgb}{0.21,0.49,0.74}
\usepackage[pagebackref,breaklinks,colorlinks,allcolors=cvprblue]{hyperref}

\def\paperID{*****} 
\def\confName{CVPR}
\def\confYear{2026}

\title{Do MLLMs Exhibit Human-like Perceptual Behaviors? HVSBench: A Benchmark for MLLM Alignment with Human Perceptual Behavior}

\author{Jiaying Lin$^{*1,3}$ \quad \quad Shuquan Ye$^{*2,3}$ \quad Dan Xu$^1$ \quad \quad Wanli Ouyang$^2$ \quad \quad Rynson W.H. Lau$^3$\\
$^1$HKUST \quad $^2$CUHK \quad $^3$City University of Hong Kong\\
}

\begin{document}

\twocolumn[{%
\renewcommand\twocolumn[1][]{#1}%
\maketitle
\vspace{-12mm}
\begin{center}
    \centering
    \captionsetup{type=figure}
    \newcommand{\teaserHeight}{5.0cm}
    \begin{minipage}[t]{0.6\textwidth}
        \centering
        \includegraphics[width=\textwidth]{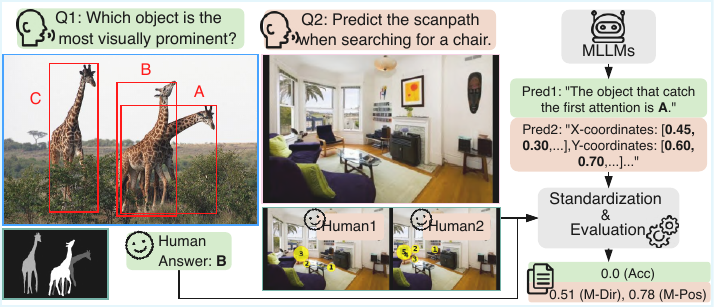}
        \vspace{-6mm}
        \caption*{(a) HVSBench\footnotemark}
        \label{fig:teaser1}
    \end{minipage}
    \begin{minipage}[t]{0.39\textwidth}
        \centering
        \includegraphics[width=\textwidth]{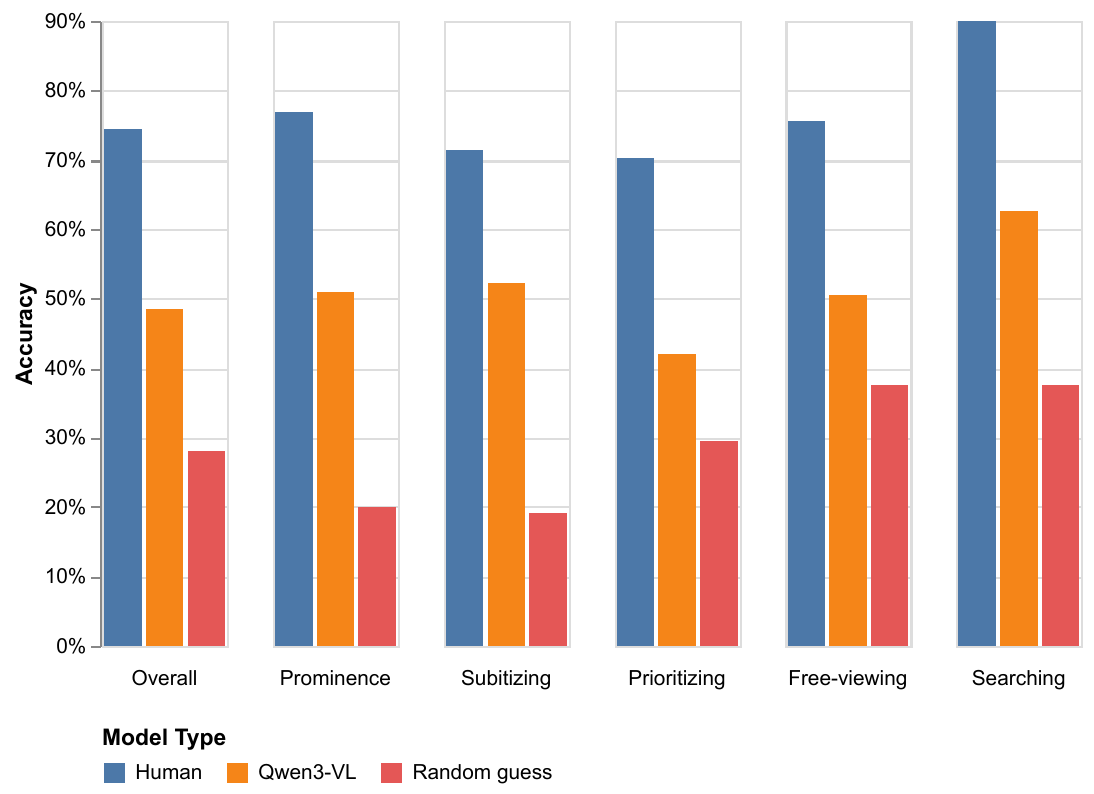}
        \vspace{-6mm}
        \caption*{(b) Results comparison. }
        \label{fig:teaser2}
    \end{minipage}
    \vspace{-2mm}
    \captionof{figure}{
    We are the first to systematically study and assess MLLMs-HVS alignment.
    (a) We propose large-scale and comprehensive HVSBench, with a robust evaluation protocol. (b) Our comparisons among humans and the state-of-the-art model Qwen3-VL on HVSBench across 5 fields reveal room for improvement and insights for developing HVS-aligned MLLMs.
    }
    \label{fig:teaser_all}
    \vspace{4mm}
\end{center}%
}]

\begin{abstract}
\begin{nolinenumbers}
\\
\vspace{-12mm}
\end{nolinenumbers}
\renewcommand{\thefootnote}{\fnsymbol{footnote}}
\footnotetext[1]{Equal contributions. Project page: \url{https://jiaying.link/HVSBench/}}

While Multimodal Large Language Models (MLLMs) excel at many vision tasks, it is unknown if they exhibit human-like perceptual behaviors. To evaluate this, we introduce HVSBench, the first large-scale benchmark with over 85,000 samples designed to test MLLM alignment with the human visual system (HVS). The benchmark covers 13 categories across 5 key fields: Prominence, Subitizing, Prioritizing, Free-Viewing, and Searching. Our comprehensive evaluation reveals a significant perceptual gap: even state-of-the-art MLLMs achieve only moderate results. In contrast, human participants demonstrate strong performance, significantly outperforming all models. This underscores the high quality of HVSBench and the need for more human-aligned AI. We believe our benchmark will be a critical tool for developing the next generation of explainable MLLMs.
\end{abstract}

\section{Introduction}
\label{sec:intro}

Recent advancements in Multimodal Large Language Models (MLLMs) have shown remarkable progress, achieving impressive performance across diverse vision-language tasks such as image captioning~\cite{coco_caption}, visual question answering~\cite{hudson2019gqa}, mathematic reasoning~\cite{lu2024mathvista}, and more. Such achievements highlight the capabilities of MLLMs in perception and vision-language interaction. 

\footnotetext{For brevity, only 2 of 10 human scanpaths are shown. Questions and predictions are simplified. Red overlayed boxes are NOT in original image.}

Despite the impressive performance of MLLMs on vision tasks, we have limited understanding of why they perform well. 
It has been demonstrated in previous studies that principles inspired by the HVS and grounded in cognitive science~\cite{desimone1995neural}, play a vital role in enhancing the performance of backbone models, such as attention-based architectures~\cite{vaswani2017attention}.
However, discrepancies remain in how MLLMs and humans perceive visual information. Human attention is based on innate and learned saliency, while MLLMs often perceive images as arrays of pixel values or feature embeddings. Human visual attention is sequential, adjusting based on context and prior knowledge, while MLLMs process input statically or through fixed-length attention. Human attention can be dynamically influenced and guided by goals, while MLLMs lack the cognition and the ability to ``refocus'' dynamically, relying purely on trained associations. 
\ysqq{Given the benefits and importance of HVS-aligned designs, including improving QA and captioning, content generation, downstream tasks, and practical applications, etc., as discussed in Sec.~\ref{subsec:discussion}
}
, the question of whether and to what extent existing MLLMs align with HVS remains a critical area for research, especially considering the current disparities.

This brings us to a fundamental question:  \textit{Do MLLMs perceive the world in the same way as humans do?} More specifically, \textit{do MLLMs fixate on similar regions of interest within an image or follow a similar temporal order as the HVS when perceiving an image? }
Humans can easily identify objects that capture their attention and perform visual searches based on context. However, in real-world scenarios, MLLMs often perceive visual information differently. 
For instance, in the first image of Fig.~\ref{fig:teaser_all}, if we ask, \textit{``Which object is the most visually prominent?''}, most people would choose the \ysqq{center giraffe, while MLLMs identify the right one}. Similarly, when searching for a chair in the second image, humans tend to first identify related objects (\eg, a table) and use contextual cues to help locate the chair (\eg, chairs are often near tables), while MLLMs may point to irrelevant areas.
However, there has been limited research and challenges on how current MLLMs align with the HVS. 
On the one hand, existing public vision-language datasets are primarily designed to assess model performance on specific tasks, offering little insight into their alignment with HVS. On the other hand, traditional HVS research has largely focused on low-level, vision-only domains~\cite{pei2022oqtr}, often evaluating models based on masks or heatmaps, making it difficult to assess MLLMs where the primary output is text rather than visual data.

In response to these challenges, we introduce a comprehensive benchmark HVSBench and designed an evaluation protocol suite for MLLMs.
Currently, HVSBench contains over 85K questions covering five distinct fields of the HVS, paired with images and answers:
\begin{enumerate}
\item 
\textbf{Prominence.} Test whether the regions MLLMs focus on align with those that are prominent to human perception. Example question in Q1 of Fig.~\ref{fig:teaser_all}.
\item \textbf{Subitizing}. Test whether the number of visually prominent objects for MLLMs matches human perception.
\item \textbf{Prioritizing.}
Assess if the order of importance assigned by MLLMs to objects reflects human viewing priorities.
\item \textbf{Free-viewing.}
Check if MLLMs can mirror the human attention shift (i.e., sequence of locations that the HSV attends to) in an image during free viewing.
\item{\textbf{Searching.}}
Test if MLLMs can follow a similar sequence of gazes as humans when searching for a specific object in an image. Example in Q2 of Fig.~\ref{fig:teaser_all}.
\end{enumerate}
The questions can be categorized into 13 types based on their phrasing, and the answer types include multiple-choice, counting (i.e., predicting an integer value), sorting, and scanpath prediction. 
To ensure both quality and variability, we design our benchmark based on a curated collection of large-scale and high-quality datasets~\cite{Yang_2020_CVPR,pei2022oqtr,chen2022characterizing,deng2024advancing} focusing on the HVS derived from real-world human studies. 
For a balanced assessment of each field, we carefully curate and divide the question field so that each field encompasses over 6,707 questions.
For the evaluation protocol, while pure exact-match metrics are unreliable due to the limitations of MLLMs in instruction-following and choice labeling, GPT-4-based matching adds bias, costs, and struggles with complex predictions like scanpaths. 
Thus, to reduce the matching-caused false-negative and improve evaluation robustness across fields, we design a human-inspired and field-adaptive automatic standardization, taking inspiration from diverse possible predictions.

We thoroughly evaluate 26 well-known SOTA MLLMs \ysqq{and human performance} on HVSBench, spanning diverse architectures and model scales. This not only provides a direct comparison among these models across multiple aspects of the HVS, but also highlights the significant gap between current MLLMs and humans. Furthermore, our findings reveal critical insights for future improvements: aligning MLLMs with the HVS cannot be achieved merely by incorporating external knowledge of related cues and priors, or by integrating human-generated captions and summaries. 
\ysqq{Additionally, we stress the value of HVS alignment across domains and their evidences: mimicking human fixation enhances QA/captioning; more intuitive content generation demonstrated by our designed prominence enhancement; improved performance in HVS-specific tasks; and applications in autonomous driving and assistive tools.}
Our contributions are summarized as follows:

\begin{enumerate}
\item To the best of our knowledge, we are \textbf{the first} to systematically study the perceptual alignment between MLLM and the HVS, across five distinct fields of HVS for model evaluation.
\item We construct HVSBench, \textbf{the first} large-scale and comprehensive benchmark with 85,147 multimodal question-answer pairs to thoroughly evaluate MLLMs in scenarios that closely mirror the HVS. 
\item We propose a robust evaluation protocol with a human-inspired and field-adaptive automatic standardization.
\item 
We conduct a comprehensive evaluation of 26 popular MLLMs \ysqq{and human} using HVSBench. Additionally, we provide new insights and techniques for developing more HVS-aligned and explainable MLLMs.
\end{enumerate}

\begin{figure*}[!ht]
    \centering
    \includegraphics[width=\linewidth]{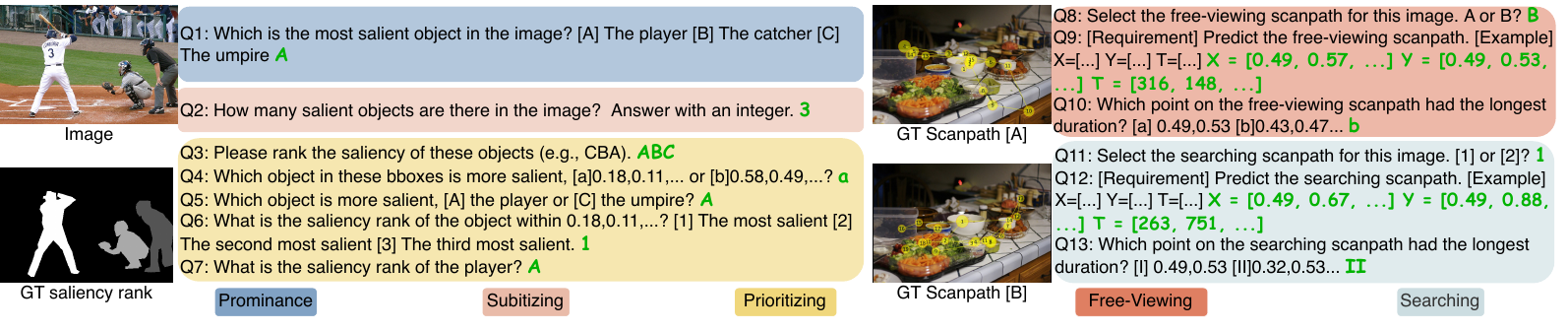}
    \vspace{-8mm}
    \caption{Simplified Samples of 13 question types in HVSBench. GT ranks and scanpath plots are for better visuals.}
    \label{fig:samples}
\end{figure*}

\section{Related Work}
\label{sec:formatting}

Existing benchmarks for MLLMs focus on assessing their capabilities in understanding and reasoning across modalities. A wide range of benchmarks, such as MMBench~\cite{liu2023mmbench} and SEEDBench~\cite{li2023seedbench}, evaluates general multimodal capabilities, including visual perception, reasoning, and comprehension. These benchmarks span diverse tasks, including document understanding~\cite{singh2019towards}, visual question answering~\cite{mathew2021docvqa}, hallucination detection~\cite{li2023evaluating} and mathematical reasoning~\cite{lu2024mathvista}. 
While existing benchmarks assess various abilities, they do not fully evaluate the alignment of MLLMs with the HVS.
The introduction of HVSBench addresses this gap, focusing on human-centric evaluation to reveal the secret of how MLLMs perceive and process visual information.
More related work on different areas is in the supplementary material.

\section{HVSBench}\label{sec:HVSBench}
We define the five distinct fields of HVS, followed by a description of multimodal QA generation, and automatic standardization and evaluation protocol.
Fig.~\ref{fig:samples} shows some examples in our HVSBench.
The Sec.~1 and Sec.~4 in supplementary materials provide the rationale behind selecting fields, their corresponding source datasets, additional samples and the results for each question type.

\subsection{Preliminaries}
To evaluate MLLM alignment with the HVS, we construct multimodal QAs based on five key fields, selected to represent diverse aspects of human visual processing and provide a comprehensive assessment. They include:

\begin{itemize}

\item \textbf{Prominence}. Identifying the most visually salient objects in a scene, reflecting human visual focus.

\item \textbf{Subitizing}. Rapidly estimating the number of salient objects, requiring parallel attention and useful for tasks like crowd analysis.

\item\textbf{Prioritizing}. Ranking objects by perceptual saliency to capture dynamic visual importance, with applications in explainable decision-making.

\item\textbf{Free-Viewing}. Modeling bottom-up gaze patterns in task-free settings with insights into visual attention.

\item\textbf{Searching}. Capturing top-down gaze behavior during goal-driven tasks, influenced by semantic context, supporting efficient, adaptive attention.

\end{itemize}
\jy{The selection of the five key fields is grounded in the dual-process theory \ysqq{introduced in~\cite{beck2009top}
}
, which distinguishes between bottom-up (stimulus-driven) and top-down (goal-directed) processes.
Bottom-up processes mainly include prominence~\cite{itti1998model}, free-viewing~\cite{jarodzka2010vector} and subitizing~\cite{trick1994small}. Top-down processes mainly include prioritizing~\cite{henderson2003human} and searching~\cite{eckstein2011visual}.
These fields were chosen because they span the majority of human visual behavior, as shown in reviews on visual scene processing~\cite{henderson2013scene}, which emphasizes saliency (Prominence), attention dynamics (Free-Viewing), and search (Searching) as critical components of HVS. While others may exist, these five fields are the most widely studied and theoretically validated in HVS to the best of our knowledge.}

\subsection{Benchmark Curation}

Based on the above definitions, we design an automatic multimodal QA generation paradigm to convert source annotations into different forms of QAs for MLLM evaluation.

\noindent\textbf{Multimodal QA Generation.}
Since the annotations in the selected datasets are not in the multiple-choice QA format, we automatically transform the ground-truth annotations into this format automatically with human verification. Specifically, we manually created 13 types of question templates. Following previous practices~\cite{liu2023mmbench}, most of our evaluations are conducted using multiple-choice QAs rather than open-ended ones, except for fields with well-defined outputs such as reporting a positive integer (\eg, subitizing) or a list of coordinates (\eg, scanpath prediction). These formats align directly with their problem metrics, enabling precise and unbiased evaluation. Open-ended answers often require scoring by LLMs or user studies, which can introduce evaluation bias or require manual intervention based on previous practices~\cite{liu2023mmbench}. Then, we create the corresponding answer options as follows.

\noindent\textbf{\ysqq{Variations in perception.}}\ysqq{We leverages \textbf{GT human} data in peer-reviewed cognitive science datasets~\cite{chen2022characterizing, deng2024advancing, pei2022oqtr, Yang_2020_CVPR} rigorously designed for diversity and bias control.
For example, Fig.~\ref{fig:teaser_all}’s source~\cite{deng2024advancing} employs 8 annotators (a common sample size~\cite{li2014secrets} in saliency research) in diverse culture (EU/CN), race (Black/White/Asian), and gender (4M/4F), collected by strict protocols (\eg, eye-tracking monitor) to minimize bias.
This ensures our benchmark captures shared human attention tendencies, aligning with common practices in HVS research.}

\noindent\textbf{(1) Template-Based Construction:} 
For most questions, we generate answer options directly from GT human annotations. 
For example, in fields like \textbf{Prominence}, the options are derived from bounding boxes or other directly annotated features. To ensure diversity in the questions, we design multiple templates for \textbf{each} question type, following the methodology outlined in~\cite{llava}. These templates provide variation while maintaining the focus of the field. For instance, in Prominence, we use templates such as:

\begin{itemize} 
\item Which object is the most salient in the image? 
\item Which object is the most visually prominent?
\item Which object attracts the highest visual attention? 
\end{itemize}

All templates are manually created by humans to ensure accuracy and relevance.

\noindent\textbf{(2) LLM-Based Refinement:} 
Note that some MLLMs may not be able to utilize the coordinate data well, but they are more suitable for interpreting in natural language. We design a bunch of questions that do not explicitly involve numerical data. However, such data cannot be easily obtained due to the limited original label types of data sources.
For scenarios where bounding box coordinates are not explicitly required and the field involves natural language references (\eg, \textit{Between these two objects — [A] \textbf{$<$obj1$>$} [B] \textbf{$<$obj2$>$} — please select the option representing the more salient object.}), we use a large language model (LLM)~\cite{achiam2023gpt4o} to generate description in natural language for the options ([A] \textbf{the person on the left side}; [B] \textbf{the bicycle on the right side}). Specifically, we use GPT-4~\cite{achiam2023gpt4o}  to describe objects based on their bounding boxes and to create plausible but false options from non-salient objects. This method ensures that the QA matches real-world applications where natural language complements visual data.

\noindent\textbf{(3) Human verification}. To address potential ambiguities introduced by LLMs or the complexity of image content in (2), human verification is employed to ensure that the referred objects in natural language descriptions correspond accurately to the intended targets. This additional step improves the accuracy and reliability of generated QAs.

\noindent\textbf{Answer Option Processing}.
For multiple-choice questions, we randomly sample answer options from the available candidates and shuffle their order. This approach enhances robustness by minimizing biases arising from fixed option orders or repetitive patterns in the answer choices.
For scanpath-related questions (i.e., free-viewing and searching), following~\cite{Yang_2020_CVPR}, we randomly sample human scanpaths from different images. This ensures that the alternative answer options are also derived from human data, making them more natural. By avoiding reliance on computer-generated scanpaths, we ensure that the evaluation remains unbiased and fair for the evaluated LLMs.

To this end, we produced 85K multimodal QAs based on annotations from 71K images, covering 13 question types across 5 key fields in HVS, including prominence, subitizing, prioritizing, free-viewing and searching. Table~\ref{tab:hvsbench_tasks} shows statistics of our HVSBench, which is large and diverse, covering various answer types and tasks.

\begin{table}[th]
\centering
\resizebox{1.0\linewidth}{!}{
\begin{tabular}{cccccc}
 \Xhline{1.0pt}
  & \makecell[c]{\textbf{Prominence} \\ Q1}  & \makecell[c]{\textbf{Subitizing} \\ Q2} &  \makecell[c]{\textbf{Prioritizing} \\ Q3-Q7} & \makecell[c]{\textbf{Free-Viewing} \\ Q8-Q10}  & \makecell[c]{\textbf{Searching} \\ Q11-Q13}  \\
 \cmidrule{2-6}
  \multirow{3}*{\makecell{\textbf{Total QAs} \\ 85,147 }}  & 8,389 (11\%) & 18,105 (24\%) & 26,309 (34\%) & 17,090 (22\%) & 6,707 (8\%)  \\
 \cmidrule{2-6}
    & \textbf{\makecell[c]{Question \\Categories}} & \textbf{\makecell[c]{Average\\ \# of Obj.}} & \textbf{\makecell[c]{Max./Avg. \\ Q Length}} & \textbf{\makecell[c]{Max./Avg. \\ A Length}} & \textbf{\makecell[c]{Max./Avg. \\ \# of Choice}} \\
 \cmidrule{2-6}
  & 13 & 3.2 & 104/45.8 & 5/1.4 & 17/3.9 \\
 \Xhline{1.0pt}
 
 \end{tabular}
 }
\vspace{-3mm}
\caption{Key statistics of HVSBench. Fixation prediction statistics are excluded due to lengthy answers.
}
\label{tab:hvsbench_tasks}
\vspace{-0.3cm}
\end{table}

\subsection{Automatic Standardization and Evaluation}
\begin{figure}[t]
    \centering
    \vspace{-3mm}
    \includegraphics[width=1\linewidth]{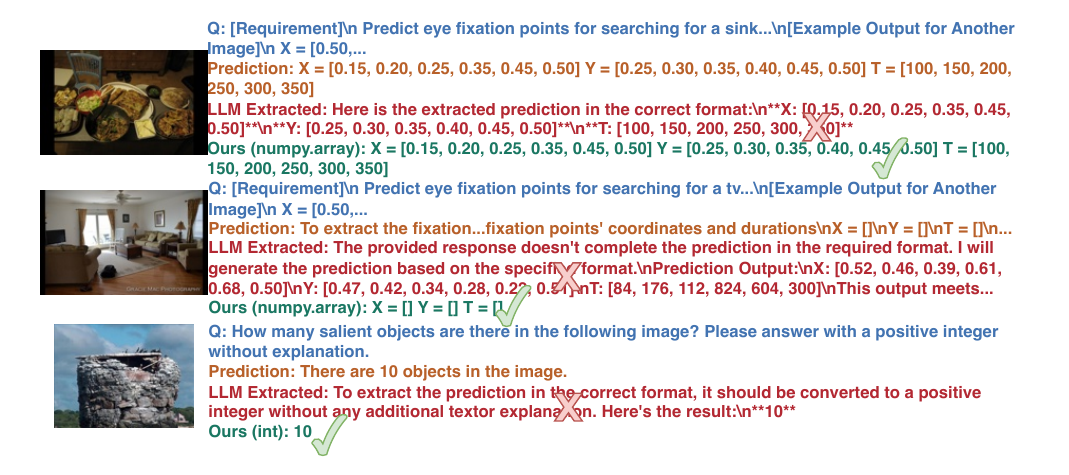}
    \vspace{-8mm}
    \caption{Illustration of our automatic standardization, which robustly formats predictions without introducing errors. In contrast, LLM-based matching (\eg, GPT-4) is both costly and prone to errors, such as failing to extract the correct format or predicting unrelated outputs.}
    \label{fig:HVSBench_evalprotocol}
\end{figure}

\noindent\textbf{Automatic Standardization}.
In evaluation protocols, pure exact-match metrics prove unreliable due to the limitations of MLLMs in instruction-following. For example, in Fig.~\ref{fig:HVSBench_evalprotocol}, even when we explicitly specify the output format and provide example outputs for reference, MLLMs still produce predictions with inconsistent and somewhat random formatting. Meanwhile, the widely adopted LLM-based matching approaches~\cite{duan2024vlmevalkit} introduce biases, incur high costs due to multiple evaluation passes, and struggle with complex predictions such as scanpaths. To address these challenges, we propose a human-inspired, field-adaptive automatic standardization method. It minimizes false negatives caused by matching errors and enhances evaluation robustness by programmatically adapting to diverse prediction formats across the five fields in our benchmark.

To achieve automatic standardization programmatically, we adopted a human-inspired, field-adaptive approach. Specifically, we observed that the responses to each question type are not open-ended but instead follow discernible patterns. Building on this insight, for each field, we randomly sampled $10$ example questions for each question type. On these example questions, we then collected multiple rounds of random predictions from all models evaluated in our experiments, as well as predictions from over $20$ different human participants and their human-annotated standardizations. With these diverse and complex predictions across fields, we programmatically defined an automated standardization process. This process ensures robust performance without errors, even on unseen predictions. 

For the \textbf{Prominence} field, the automatic standardization cleans and normalizes predictions into consistent main choice labels (\eg, `A', `C', etc.) from various input formats, using regex to detect explicit or implied labels in diverse phrasing. 
For \textbf{Subitizing}, the automatic standardization processes predictions to extract integer values, including direct numbers, text-based numbers (\eg, ``three''), or implied quantities. When no valid number is found, it defaults to the average ground truth value in 3.2116. 
For \textbf{Prioritizing}, except for choice answers that are processed the same as in the Prominence field, our automatic standardization detects and normalizes answers of sequence (sorting) via order-related phrasing through regex patterns.
In \textbf{Free-viewing} and \textbf{Searching} fields, The automatic standardization involves extracting and normalizing scanpath data (X, Y for coordinates and T for durations) from diverse input formats, including text, JSON, and various phrasing styles. It also addresses irregularities like mismatched list lengths and incomplete scanpaths.
Refer to Sec. 2 in supplementary for predictions and their corresponding standardized outputs in each field.
We illustrate the superiority of our automatic standardization over LLM-based matching in Fig.~\ref{fig:HVSBench_evalprotocol}. In our experiments, GPT-4 was prompted with the full question, the full prediction, and the text prompt: ``Please extract the prediction with the correct format.'' Refer to Sec. 3 in supplementary for detailed inputs, experimental settings, and pseudo-code for our automatic standardization.

\noindent\textbf{Evaluation Metrics}.
In the \textbf{Prominence} field, we use accuracy in which are choice questions.
For the \textbf{Subitizing} field, we adopt widely used metrics, including Mean Absolute Error (MAE), Root Mean Square Error (RMSE), as well as Exact Match Accuracy (Acc). 
For \textbf{Prioritizing}, the answer types include single-choice and sequence (sorting). The single choices are evaluated via accuracy. To evaluate whether the predicted order matches the ground truth exactly, Exact Match Accuracy (Acc) is also used.
In \textbf{Free-viewing} and \textbf{Searching}, we employ the widely adaptive MultiMatch~\cite{jarodzka2010vector} similarity to evaluate generated scanpaths across shape, direction, length, and position. Our main results focus on the direction ("M-Dir") and position ("M-Pos") for simplicity. 
Following standard procedures~\cite{Yang_2020_CVPR}, predictions are cropped to length 6, with final scores for each prediction being MultiMatch averaged over 10 GT scanpaths in the image.

\begin{table*}[!h]
    \centering
    \resizebox{0.99\linewidth}{!}{
 \begin{tabular}{lcccccccccccc}
\Xhline{1.0pt}
\multirow{3}{*}{Models} & \multirowcell{3}{Overall \\ Acc$\uparrow$} & Prominence  & \multicolumn{3}{c}{Subitizing}  & Prioritizing   & \multicolumn{3}{c}{Free-viewing}     & \multicolumn{3}{c}{Searching}    \\
          & & Q1 & Q2 & Q2 & Q2 & Q3-Q7 & Q8,Q10 & Q9 & Q9 & Q11,Q13 & Q12 & Q12 \\

          &         &    Acc$\uparrow$ &    Acc$\uparrow$ &     MAE$\downarrow$ &         RMSE$\downarrow$ &            Acc$\uparrow$ &            Acc$\uparrow$ & M-Dir$\uparrow$ & M-Pos$\uparrow$ &         Acc$\uparrow$ & M-Dir$\uparrow$ & M-Pos$\uparrow$ \\
\Xhline{0.8pt}
\rowcolor{LGray} \multicolumn{13}{c}{\textbf{Human Performance}} \\
Human &  0.7438 &  0.7683 & 0.7143 & 0.6349 & 1.3333 &         0.7025 &          0.7560 & 0.6963 & 0.8863 &      0.9000 & 0.7617 & 0.9181 \\
\hline
\rowcolor{LightGreen} \multicolumn{13}{c}{\textbf{Baselines}} \\
Random guess &  0.2806 &  0.1994 & 0.1909 & - & - &         0.2949 &          0.3750 & - & - &   0.3752 & - & - \\
\hline
\rowcolor{LightYellow} \multicolumn{13}{c}{\textbf{Proprietary MLLMs}} \\ 
GPT4-o &  0.3946 &  0.3139 &   0.4512 &    1.3445 &       3.2614 &         0.3621 &         0.3737 & 0.5917 & 0.8042 &      0.4015 & 0.5106 & 0.7834 \\
GPT5-nano & 0.4435 & 0.4664 & 0.4931 & 1.1840 & 2.2897 & 0.4117 & 0.3939 & 0.5961 & 0.8143 & 0.5885 & 0.4884 & 0.8004 \\
GPT5-mini & 0.4675 & 0.4896 & 0.5493 & 1.0080 & 2.0429 & 0.4219 & 0.3969 & 0.6088 & 0.8100 & 0.6509 & 0.6150 & 0.8394 \\
Gemini-1.5-Flash &  0.3886 &  0.3323 &  0.5106 &    1.3070 &       3.0978 &         0.3283 &         0.3804 & 0.6128 & 0.8392 &      0.4040 & 0.5199 & 0.8206 \\
Gemini-2.0-flash & 0.4049 & 0.4652 & 0.4602 & 1.9602 & 23.3327 & 0.3839 & 0.3588 & 0.5739 & 0.8239 & 0.3616 & 0.4876 & 0.8358 \\
Gemini-2.0-flash-lite & 0.3556 & 0.3932 & 0.3208 & 2.0530 & 4.4785 & 0.3495 & 0.3778 & 0.6085 & 0.7996 & 0.3915 & 0.5411 & 0.8170 \\
Gemini-2.5-flash & 0.4221 & 0.4408 & 0.3865 & 2.2709 & 5.9720 & 0.4215 & 0.4142 & 0.6298 & 0.8393 & 0.5885 & 0.6252 & 0.8540 \\
Gemini-2.5-flash-no-thinking & 0.4148 & 0.4774 & 0.4321 & 2.0069 & 4.9635 & 0.4061 & 0.3778 & 0.6165 & 0.8481 & 0.4214 & 0.5221 & 0.8202 \\
Gemini-2.5-flash-lite & 0.4071 & 0.4457 & 0.4608 & 1.3977 & 4.1724 & 0.3720 & 0.3915 & 0.5640 & 0.8460 & 0.3840 & 0.4693 & 0.8361 \\
Gemini-2.5-pro & 0.4754 & 0.5189 & 0.5308 & 1.1501 & 2.5422 & 0.4505 & 0.4082 & 0.6251 & 0.8396 & 0.5810 & 0.6378 & 0.8583 \\
\hline
\rowcolor{LightBlue} \multicolumn{13}{c}{\textbf{OpenSource MLLMs}} \\ 
DeepSeek-VL  &   0.3655 & 0.3223 &   0.4544 &    1.2471 &       2.2514 &         0.3327 &         0.3445 & 0.5118 & 0.5450 &      0.3516 & 0.4790 & 0.7130 \\
Idefics2 &   0.3067 & 0.2015 &  0.2990 &    2.2078 &      3.8573 &          0.2971 &          0.3637 & 0.5151 & 0.5525 &      0.3865 & 0.5322 & 0.7619 \\
Idefics3  &   0.3552 & 0.2149 &  0.4852 &    1.1055 &      1.9567 &          0.3272 &          0.3411 & 0.5329 & 0.6310 &      0.2843 & 0.5407 & 0.6856 \\
LLaVA-Next  &  0.3460 &  0.3223 &    0.3961 &   1.3727 &        2.2675 &         0.3341 &         0.3274 & 0.5249 & 0.5998 &      0.3192 & 0.5083 & 0.7883 \\ 
LLaVA-OneVision  &   0.4035 & 0.4640 &   0.4517 &    1.1206 &        1.9671 &         0.3495 &         0.3888 & 0.5490 & 0.8177 &      0.4913 & 0.5047 & 0.7295 \\ 
mPLUG-Owl3  &  0.3076 &  0.3309 &   0.2688 &    2.8941 &       4.6162 &         0.3305 &         0.3117 & 0.5028 & 0.4941 &      0.2668 & 0.5368 & 0.5592 \\
MiniCPM-V 2.6  &  0.3476 &  0.3748 &  0.4491 &   1.2471 &       2.1137 &         0.2989 &         0.3023 & 0.5132 & 0.5589 &      0.3416 & 0.4953 & 0.7107 \\
ERNIE-4.5-VL-28B-A3B & 0.3936 & 0.4115 & 0.4496 & 1.5673 & 3.7682 & 0.3599 & 0.3766 & 0.5562 & 0.8495 & 0.3990 & 0.5120 & 0.7913 \\
InternVL2.0 &  0.3082 &  0.3101 &  0.3405 &   2.3458 &       3.6068 &         0.3405 &         0.3409 & 0.5028 & 0.7090 &      0.2968 & 0.5016 & 0.7282\\
InternVL3.5-8B & 0.4254 & 0.4811 & 0.5111 & 1.1426 & 2.1146 & 0.3839 & 0.3564 & 0.5013 & 0.6447 & 0.4863 & 0.4683 & 0.7919 \\
GLM-4.1V-9B-Base & 0.3932 & 0.4664 & 0.4040 & 1.3828 & 2.3392 & 0.3961 & 0.3474 & 0.5139 & 0.7772 & 0.3641 & 0.4976 & 0.7625 \\
GLM-4.1V-9B-Thinking & 0.4191 & 0.4872 & 0.3834 & 1.9067 & 3.6251 & 0.4057 & 0.4195 & 0.6022 & 0.7902 & 0.5387 & 0.5113 & 0.8351 \\
GLM-4.5V & 0.4640 & 0.4737 & 0.5286 & 1.2582 & 2.6779 & 0.4380 & 0.3999 & 0.5982 & 0.7756 & 0.5885 & 0.5343 & 0.8096 \\
Qwen2-VL-7B  &   0.4076 & 0.4103 &  0.5090 &     1.4698 &       2.6747 &         0.3901 &         0.3182 & 0.5199 & 0.7178 &      0.4214 & 0.4610 & 0.8200 \\
Qwen3-VL-8B & 0.4320 & 0.4811 & 0.4539 & 1.1363 & 1.9546 & 0.4090 & 0.4011 & 0.5122 & 0.8421 & 0.5187 & 0.4812 & 0.8613 \\
Qwen3-VL-235B-A22B & 0.4851 & 0.5092 & 0.5223 & 1.0483 & 1.9148 & 0.4204 & 0.5054 & 0.5665 & 0.8468 & 0.6259 & 0.5109 & 0.8517 \\
\Xhline{1.0pt}
\end{tabular}
}
    \vspace{-0.3cm}
    \caption{\textbf{HVSBench Leaderboard.} The results of MLLMs reveal significant room for improvement. 
    }
    \label{tab:leaderboard}
    \vspace{-0.4cm}
\end{table*}

\section{Experiments}

\subsection{Evaluation Details}
In Subitizing of random guess, accuracy is computed using random sampling based on the frequency distribution of subitized object counts. 
\ysqq{Our human evaluation involves 10 participants from diverse backgrounds, considering individual differences~\cite{koehler2014saliency}. For MultiMatch, we use a commonly used oracle method~\cite{Yang_2020_CVPR}\jy{, which compares each subject's scanpath as a prediction to others' as GT and averages the results.}
}
We select recent open-source MLLMs for validation, including DeepSeek-VL~\cite{lu2024deepseekvl}, Idefics-series~\cite{laurencon2023obelics_idefics,laurenccon2024building} LLaVA-Next~\cite{llava}, LLaVA-OneVision (LLaVA-OV)~\cite{li2024llava_onevision}, mPLUG-Owl3~\cite{ye2024mplug}, Qwen-VL series~\cite{wang2024qwen2vl},  InternVL-series~\cite{chen2024far_internvl2}, ERNIE-4.5-VL~\cite{ernie2025technicalreport}, MiniCPM-V 2.6~\cite{yao2024minicpm} and GLM-4V series~\cite{vteam2025glm45vglm41vthinkingversatilemultimodal}. We evaluate a range of model scales, from 7B up to 235B. Note that we still focus on 7B-scale models for both efficiency, effectiveness and fair comparisons among baselines by selecting most baselines at this scale. We also include GPT series~\cite{achiam2023gpt4o} and Gemini series~\cite{team2024gemini}, two representative proprietary MLLMs, as baselines in our evaluation. 
All experiments are conducted using VLMEvalKit~\cite{duan2024vlmevalkit} on the same platform for consistency and fairness. 
\ysqq{We allocate 10\% of HVSBench for evaluation and reserve the remaining 90\% for further explorations such as instruction-tuning MLLMs.}

\begin{figure*}[htp]
    \centering
    \includegraphics[width=0.95\linewidth]{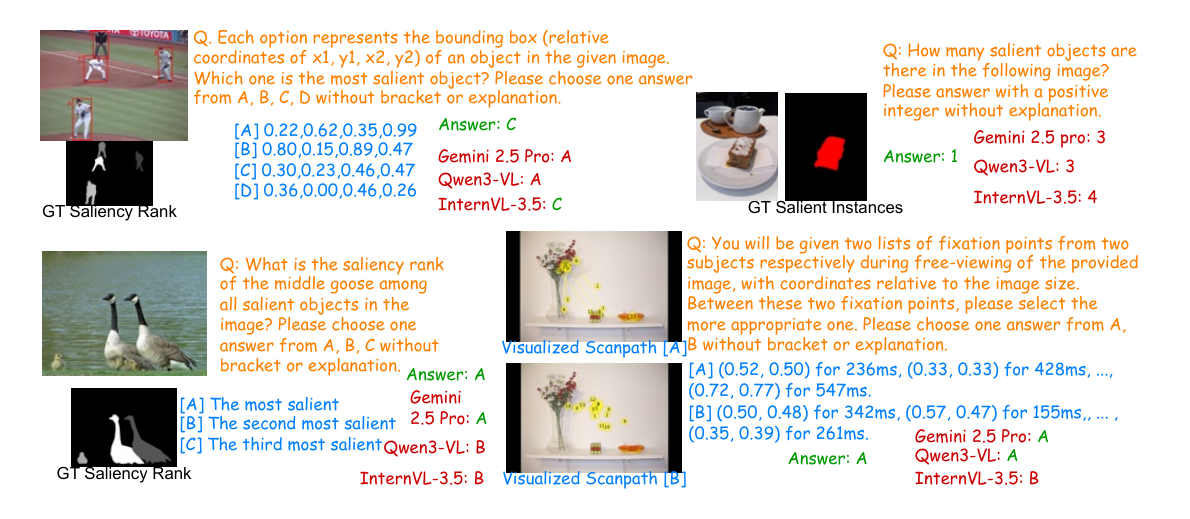}
    \vspace{-5.5mm}
    \caption{Qualitative results. The bounding boxes, the scanpaths and the GT masks from the source datasets (\eg, rank \& instances) are 
    for visual clarity and not used in the input images for evaluation. 
    Text is partially omitted due to limited space.
    }
    \label{fig:qualitative}
\end{figure*}

\subsection{Main Results on HVSBench}

\noindent \textbf{Quantitative Evaluation.}
Table~\ref{tab:leaderboard} presents the evaluation results on HVSBench, highlighting that current open-source and proprietary MLLMs still underperform in aligning with HVS. \ysqq{Diverse human participants achieved strong performance (OA: 0.7438).} \ysqq{Given individual differences in saliency and cognitive science~\cite{koehler2014saliency}, this performance is both high and reasonable, which further validates the quality of our benchmark.} \ysqq{Humans (OA: 0.7438) clearly outperform both random guessing (OA: 0.2806) and the best MLLMs (best OA: 0.4851)  in all metrics.} \ysqq{The top-performing MLLMs are the open-source Qwen3-VL-235B (OA: 0.4851) and the proprietary Gemini-2.5-pro (OA: 0.4754). Notably, the best open-source model slightly outperforms the best proprietary model. Both models significantly surpass older models like GPT4-o (OA: 0.3946), demonstrating the rapid potential of recent MLLMs.}

In Free-viewing (Acc Q8,Q10), several models now outperform random guessing (0.3750), including LLaVA-OneVision (0.3888), Gemini-2.5-pro (0.4082), and the top-performing Qwen3-VL-235B (0.5054). The performance is even stronger in Searching (Acc Q11,Q13), where numerous models like LLaVA-OneVision (0.4913), Gemini-2.5-pro (0.5810), and Qwen3-VL-235B (0.6259) significantly surpass the random guess baseline (0.3752). A possible explanation is that in Free-viewing, which is a pure vision field that reflects HVS behavior without conditioning, these models exhibit behaviors that differ from humans. In contrast, Searching involves more defined objectives and patterns that are uniform and predictable, making them easier for MLLMs to align with.

Despite this progress, the gap to human performance remains. These results suggest significant room for scanpath prediction for current MLLMs. The Sec.~6 in supplementary materials shows the detailed benchmark on all 13 question types.

\noindent \textbf{Benchmark Examples and Predictions.}
Fig.~\ref{fig:qualitative} shows examples and results of the three best-performed and representative models. We include the GT saliency information based on natural viewing by human observers for reference. 
The top-left illustrates a sample from the Prominence field. A human can easily identify the person closest to the center as the most salient. However, both Gemini 2.5 Pro and Qwen3 fail in this case. 
The top-right corner shows a sample from the Subitizing field. All baselines fail to predict the correct quantity. 
The bottom-left corner features a sample from the Prioritizing field, where only Gemini 2.5 Pro succeeds. 
The bottom-right corner displays a sample from the Free-Viewing field. Option A reflects a clear pattern of human attention, while the human scanpath shown in option B is unrelated to the image. Despite this, the InternVL-3.5 incorrectly selected B. 
The bottom-right shows a surprising case in our HVSBench where advanced MLLMs can select the human free-viewing scanpath correctly, even though such a task is purely vision and not included in their training objectives. This highlights that the advanced MLLMs behavior in purely vision tasks may have some similar properties like those in HVS. 
Refer to Sec. 4 in the supplementary for more qualitative results.

\noindent\textbf{Do thinking models outperform non-thinking models?}.
Recent MLLMs show outperformance by utilizing thinking processes. To investigate this, we manually disable the thinking process of gemini-2.5-flash  (``gemini-2.5-flash-no-thinking'') and found that its non-thinking model generally performs worse than the origin thinking model. In particular, the overall accuracy drops from 0.4221 to 0.4148, while we observe a significant drop in the searching section (Acc: 0.5885$\rightarrow$0.4214). This is reasonable since visual search requires contextual modeling during its searching process and a thinking-based model is good at it. 

\noindent\textbf{MLLMs' Explanation on its Choice}.
Taking the bottom left image used in Fig.~\ref{fig:qualitative} as an example, Gemini 2.5 Pro can explain why it predicts the middle goose is the most salient:
\textit{Size: It is a large subject in the image.
Contrast: Its long, black neck and distinct white cheek patches create a strong visual contrast against its own body and the muted green water in the background.
Position: It is centrally located and in sharp focus, acting as the primary focal point of the image.}  
This explanation shares the findings from cognitive science on HVS~\cite{itti1998model}, and it shows advanced MLLMs like Gemini 2.5 Pro can somehow align with HVS and understand the concept of saliency.

\begin{figure}[h]
    \centering
    \captionsetup[sub]{labelformat=empty,skip=0pt} 
    \begin{subfigure}[b]{0.45\linewidth}
        \includegraphics[width=\linewidth]{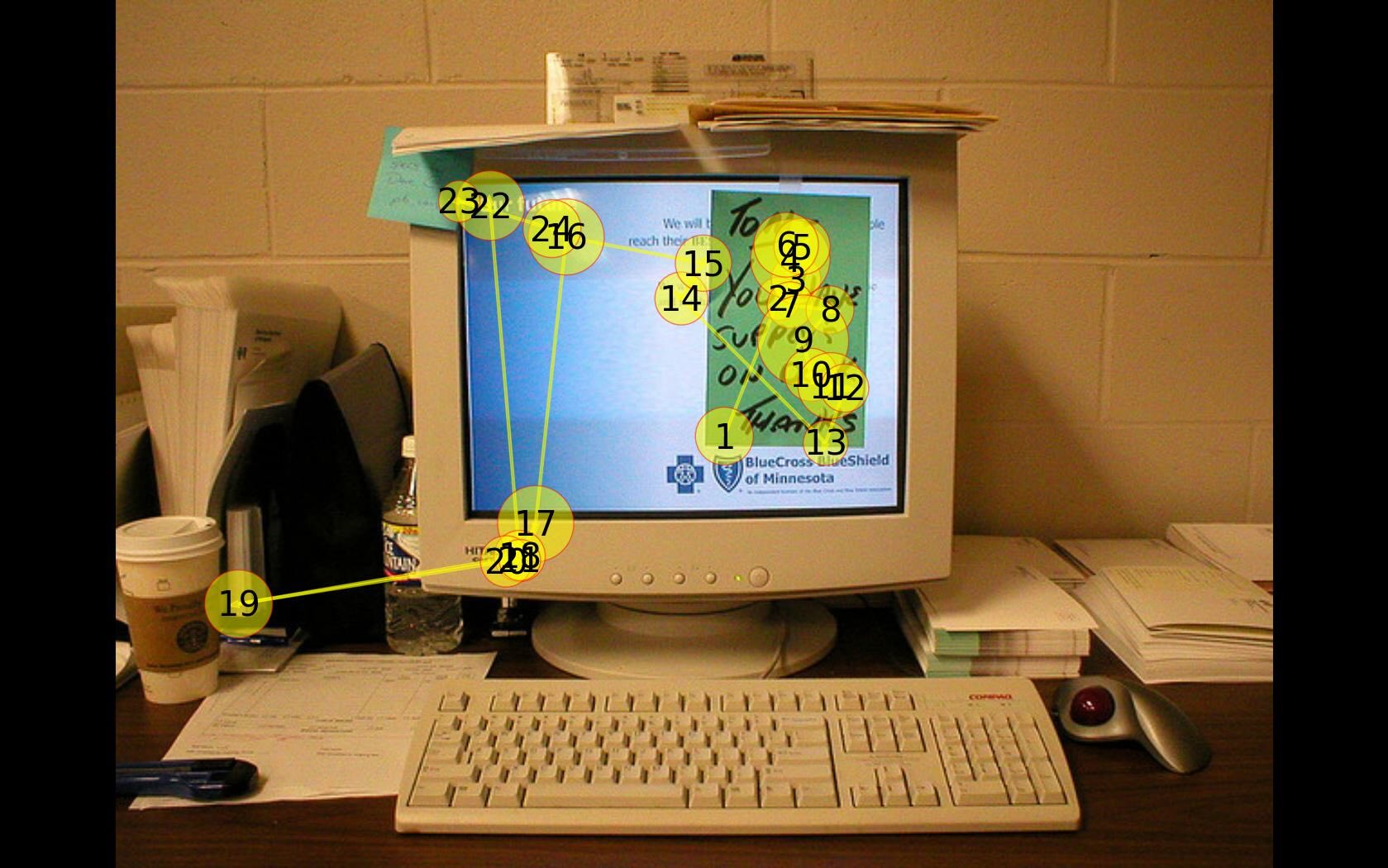}
        \caption{GT (Human)}
        \label{fig:gt}
    \end{subfigure}
    \begin{subfigure}[b]{0.45\linewidth}
        \includegraphics[width=\linewidth]{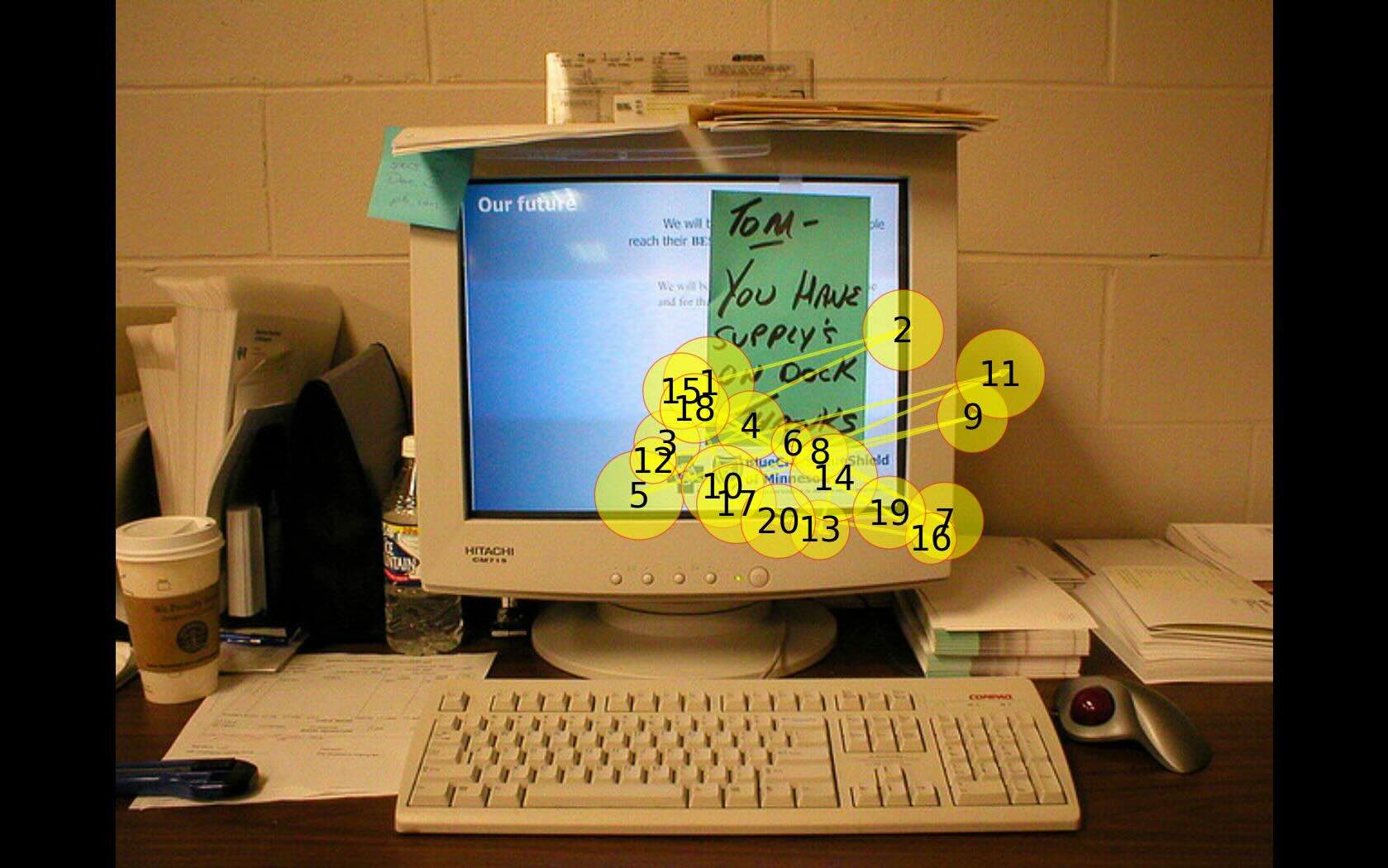}
        \caption{MLLM Prediction}
        \label{fig:gemini}
    \end{subfigure}
    \vspace{-3.5mm}
    \caption{MLLMs' predictions differ from GT reasonably. MLLMs and GT differ clearly from random guess.}
    \label{fig:comparison}
\end{figure}

\noindent\textbf{MLLMs' Scanpaths v.s. Human's GT}. In Fig.~\ref{fig:comparison} shows that the GT scanpaths obtaining from humans focus on text and informative areas, while MLLMs' scanpaths are generally reasonable (focus on objects), slightly differing from humans (text first). This shows MLLMs share similar observing pattern with humans but still have a significant gap, as shown in Table~\ref{tab:leaderboard}.

\subsection{Ablation Study}

\noindent\textbf{Model Size.} We assess model size impact on HVSBench by testing models with different parameter counts. We select two representative methods: DeepSeek-VL~\cite{lu2024deepseekvl} and GPT-4o~\cite{achiam2023gpt4o}, from both open-source and proprietary MLLMs, for comprehensive analysis. As shown in Table~\ref{tab:ablation_params}, larger MLLMs generally outperform smaller ones across all metrics. It suggests that increasing model size leads to better alignment with HVS for MLLMs. \jy{Refer to the supplementary for more experiments.}

\begin{table}[t]
    \centering
    \resizebox{\linewidth}{!}{
\begin{tabular}{lcccccc}
        \Xhline{1.0pt}

 Baselines & \# Param &  PO $\uparrow$ & \multicolumn{1}{c}{SU$\uparrow$}  &     PI $\uparrow$  & \multicolumn{1}{c}{FV$\uparrow$}     & \multicolumn{1}{c}{SE$\uparrow$}    \\
        \Xhline{0.8pt}
GPT4-o mini~ &   N/A & 0.3126   & 0.4480 &       0.3312 &         0.3560  &      0.3766 \\ 
GPT4-o  & N/A   & \textbf{0.3139}  &   \textbf{0.4512} &              \textbf{0.3621} &         \textbf{0.3737}  &      \textbf{0.4015} \\ \hline
\multirow{2}{*}{DeepSeek-VL} &  1.3B & 0.1758   & 0.2513 &       0.2950 &         0.3188  &      0.2843 \\
   & 7B  & \textbf{0.3223}   &    \textbf{0.4544} &              \textbf{0.3327} &         \textbf{0.3445}  &      \textbf{0.3516} \\ \hline
   
\Xhline{1.0pt}
\end{tabular}
}
    \vspace{-0.3cm}
    \caption{\textbf{Ablation study of the number of params.} PO, SU, PI, FV and SE  means ``Prominence'', ``Subitizing'', ``Prioritizing'', ``Free-viewing'', ``Searching'', respectively. 
    }
    \label{tab:ablation_params}
    \vspace{-0.4cm}
\end{table}

\noindent\textbf{Human Captions or Descriptions Improve Alignment?} 
It is possible that certain annotations, like human captions \jy{from COCO caption} and detailed descriptions \jy{from LLaVA-Instruct-150K}, implicitly reflect the HVS. Therefore, we evaluated whether adding detailed descriptions (``Detail.'') or short captions (``Cap.'') could improve the performance of the base GPT-4o model. 
As shown in Table~\ref{tab:ablation_question}, in the field of Prominence and Subitizing, adding detailed descriptions and captions improved performance, suggesting that additional information helps the model better identify and subitize visually dominant elements. 
In Prioritizing, while detailed descriptions provided some improvements, captioned input worsened the score, suggesting that brief caption information may not be sufficient for understanding object importance in a human-aligned way. In Free-viewing, both detailed descriptions and captions actually lowered performance. These results suggest that neither detailed descriptions nor captions can effectively capture human attention shifts in free-viewing. In Searching, captions slightly improve accuracy, while detailed descriptions reduce accuracy. 

One reason may be that humans typically shift their attention rapidly between the most salient objects when searching. Captions, by prioritizing only the most salient features, better mirror this fast-paced, targeted attention. Detailed descriptions, however, broaden the focus range, leading to attention that doesn’t align with human searching. 
These results suggest that providing more context through human captions or detailed descriptions can lead to better performance in some evaluation criteria, aligning MLLMs with the HVS cannot be achieved merely by integrating human-generated captions and summaries, especially in Prioritizing, Free-viewing and Searching fields.


 
      


\begin{table}[t]
    \centering
    \resizebox{\linewidth}{!}{
\begin{tabular}{lcccccc}
        \Xhline{1.0pt}

 Baselines & Hint & PO $\uparrow$ & \multicolumn{1}{c}{SU$\uparrow$}  &     PI $\uparrow$  & \multicolumn{1}{c}{FV$\uparrow$}     & \multicolumn{1}{c}{SE$\uparrow$}    \\
        \Xhline{0.8pt}
 
\multirow{3}{*}{GPT-4o} &  \xmark & 0.3139  &   0.4512 &              0.3621 &         \textbf{0.3737}  &      0.4015 \\
      
 &  Detail. & \textbf{0.4274} &      0.4809 &              \textbf{0.3799} &         0.3465  &      0.3791 \\
    & Cap. & 0.4188 &     \textbf{0.4852} &              0.3464 &         0.3605  &      \textbf{0.4190} \\ \hline

\Xhline{1.0pt}
\end{tabular}
}
    \vspace{-0.3cm}
    \caption{Ablation study of question prompt.}
    \label{tab:ablation_question}
    \vspace{-0.4cm}
\end{table}

\noindent\textbf{Field-Specific Hints Improve Alignment?} 
Since definitions and prior knowledge of the fields also provide cues for the HVS, we examine whether adding Field-Specific Hints can enhance model performance. For example, the simplified hint for the Prominence field is:
\textit{The detection of salient objects aims to simulate the human visual perception system by identifying and localizing the most visually striking object(s) in a scene. ... (omitted), cues such as color contrast, spatial bias, and depth contrast also influence saliency.} 
Refer to Sec. 5 in supplementary for full hints in each field.

Table~\ref{tab:ablation_system} summarizes the results on GPT-4o. Our results show an improvement in the Prominence, Subitizing, and Prioritizing fields but a noticeable decrease in performance for Free-viewing and Searching. This decline suggests that too much contextual information may hinder the model’s focus on the raw visual features necessary for free-viewing tasks. A possible explanation is that prior knowledge might cause the model to focus immediately on specific parts of the image, disrupting the natural temporal sequence of the human gaze. This could lead to fixation sequences that are less representative of human free-viewing or searching.

\begin{table}[t]
    \centering
    \resizebox{\linewidth}{!}{
\begin{tabular}{lcccccc}
        \Xhline{1.0pt}

 Baselines & Task & PO $\uparrow$ & \multicolumn{1}{c}{SU$\uparrow$}  &     PI $\uparrow$  & \multicolumn{1}{c}{FV$\uparrow$}     & \multicolumn{1}{c}{SE$\uparrow$}    \\
        \Xhline{0.8pt}
\multirow{2}{*}{GPT-4o} &  \xmark & 0.3139 &    0.4512 &              0.3621 &         \textbf{0.3737}  &      \textbf{0.4015} \\
    & \cmark & \textbf{0.4212} &    \textbf{0.5005}   &          \textbf{0.3803} &         0.3476  &      0.3342 \\ \hline

\Xhline{1.0pt}
\end{tabular}
}
    \vspace{-0.3cm}
    \caption{Ablation study of system prompt.}
    \label{tab:ablation_system}
\end{table}

\begin{table}[t]
    \centering
\begin{tabular}{lc}
        \Xhline{1.0pt}

 Baselines & Search Length $\downarrow$    \\
        \Xhline{0.8pt}
Random-DFS & 9.97 \\
Human Fixation & \textbf{2.52} \\
\Xhline{1.0pt}
\end{tabular}
    \vspace{-0.3cm}
    \caption{Human fixation can guide the visual search process in \cite{wu2024v} on COCO-Search18, obtaining from~\cite{wu2024v} with simplication.}
    \label{tab:vstar}
    \vspace{-0.4cm}
\end{table}
\subsection{\ysqq{Discussion: Benefits and Applications}}\label{subsec:discussion}

\ysqq{While human perception mimicry may not be universally beneficial, but we stress its value in key tasks.}
\textbf{(1)} Mimicking human perception can significantly improve tasks requiring precise visual grounding, such as QA, captioning and visual searching.
As shown in Table~\ref{tab:vstar} (which is obtained from the \ysqq{Table 4 of}~\cite{wu2024v}), models mimicking human fixation mechanisms achieve shorter search lengths on COCO-Search18, improving efficiency and accuracy. 
\ysqq{\textbf{(2)} In HVS-related downstream tasks, models better aligned with HVS perform better. }
\ysqq{For example,
Qwen2-VL outperforms MiniCPM-V in our HVSBench and also in general tasks such as human perception, visual illusion (MMMU, HallusionBench)
, but is comparable or worse in tasks unrelated to HVS like math reasoning (MathVision).}
\textbf{(3)} Practical applications like autonomous driving, perception in robots, assistive tools for the visually impaired, and tasks directly related to the HVS (\eg, saliency prediction),
further demonstrate the benefits where HVS alignment ensures intuitive, user-centered outputs.
\textbf{(4)} We also provide a potential application and show that content generation models better aligned with the HVS can produce more reasonable outputs.
\ysqq{Taking the prominent field as an example, we design a smart thumbnailing based on cropping and prominence enhancement to illustrate as shown in Fig.~\ref{fig:application_crop}.}
Specifically, we examine how GPT-4o crops the image to enhance the prominence of one object: a photo. 
GPT-4o with a task-specific hint generates a reasonable analysis and successfully crops the image to highlight the photo, compared to the result without hint, demonstrating better alignment with HVS.
This can be directly applied to automated design, context-aware content generation, and visual storytelling.

\begin{figure}[t]
    \centering
    \includegraphics[width=0.9\linewidth]{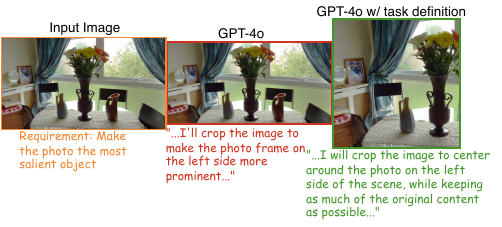}
    \caption{Application: Smart Thumbnailing. 
    Based on the results from Table 5 (PO: w/o task definition 0.3139 $\rightarrow$ w/ task definition 0.4274), GPT-4o with better performance on prominence (i.e., w/ task definition, PO: 0.4274) can generate a better smart thumbnail than the one w/o task definition (PO: 0.3139).}
    \label{fig:application_crop}
\end{figure}

Based on these aspects, we believe mimicking human perception will benefit machine vision.

\section{Conclusion}
In this paper, we explore the alignment between MLLMs and the human visual system. We introduce HVSBench, a novel large-scale benchmark designed to evaluate MLLMs on vision tasks that closely mirror human perception. It consists of 85K multimodal QAs across 13 categories and 5 fields, accompanied by a robust evaluation protocol.
Our experiments demonstrate that HVSBench poses a new, significant challenge for state-of-the-art MLLMs, highlighting considerable room for improvement. We believe HVSBench will drive the development of more human-aligned and explainable MLLMs, offering critical insights into how these models perceive and process visual information.

\clearpage
\setcounter{page}{1}
\maketitlesupplementary

\part*{}

The supplementary materials include:
\begin{enumerate}
    \item \textbf{Section 1: Preliminary Settings for HVBench} \\
    This section explains the rationale behind the selection of fields for HVBench. It also provides details about the corresponding source datasets used for each selected field.
    
    \item \textbf{Section 2: Sample Prediction and Standardization} \\
    We present the predictions generated by the models and the corresponding outputs after applying our standardization process for each field. This ensures consistency and comparability across different fields.
    
    \item \textbf{Section 3: Detailed Settings} \\
    This section provides detailed settings in our experiments and standardization pipeline, including the inputs used in our experiments, the detailed experimental settings, and the pseudo-code for our automated standardization pipeline.
    
    \item \textbf{Section 4: Additional Qualitative Examples and Results} \\
    We include extra qualitative examples showcasing sample questions and the predictions made by MLLMs under each field. 
    
    \item \textbf{Section 5: Field-Specific Hints in Each Field} \\
    For each field, we provide specific hints or context that help understand tasks.
    
    \item \textbf{Section 6: Detailed Benchmark Results and Disucussions} \\
    This section contains a detailed report of benchmark results for all 13 question types, offering a comprehensive view of the performance across various fields and question types.
    We also provide additional ablation studies and related discussions for them.
    
    \item \textbf{Section 7: More Related Work} \\
    We include more related work for different research areas to showcase the necessity of our HVSBench.
    \item \textbf{Section 8: Application: } \\
    We include an application, which shows that content generation models better aligned with the HVS can produce more reasonable outputs.
\end{enumerate}

\section{Preliminary setting for HVBench}
\label{sec:rationale}

To evaluate the extent to which multimodal large language models (MLLMs) align with the HVS, we construct multimodal queries based on five fields designed for different datasets. These fields are critical for assessing whether MLLMs perceive and interpret visual information in a manner akin to humans. For a fair and comprehensive evaluation, the five fields are carefully curated to capture different aspects of the HVS. These fields include:

\noindent \textbf{Prominence}.
The prominence in HVS enables humans to identify the most visually prominent objects within an image~\cite{pei2022oqtr}, making it a critical application for understanding human visual focus in the HVS.  
In this field, we choose the SIFR dataset~\cite{deng2024advancing} to construct our benchmark data. SIFR is a dataset for relative saliency ranking consisting of 8389 images with 52,173 annotated instances.

\noindent \textbf{Subitizing}. 
Subitizing~\cite{zhang2015salient} is to quickly and accurately perceive the number of visually prominent objects in a scene. Compared to Prominence, it requires simultaneous attention to multiple elements. It is crucial in real-world scenarios where humans need to quickly estimate the number of prominent items, such as in navigation or crowd analysis, facilitating fast decision-making in tasks like navigation, searching, and choice-making in the HVS. 
We choose SIFR dataset~\cite{deng2024advancing} and SIS10K~\cite{pei2022oqtr} for this field since the original subitizing dataset~\cite{zhang2015salient} is no longer available. SIS10K is a large-scale salient instance segmentation dataset. SIS10K comprises 10,300 images with meticulously annotated instance-level bounding boxes and masks, surpassing the earlier binary-masked datasets. Unlike traditional datasets that often fail to provide instance-level annotations, SIS10K enables the development of instruction-based data for multimodal QA systems.  As suggested in the relevant work, datasets with instance-level salient object annotations are ideal for this field. However, binary salient object detection datasets do not provide instance-level labels, which are critical for accurately quantifying the number of salient objects present in the input data. This limitation highlights the importance of using datasets that explicitly support instance-level annotations to ensure reliable performance in subitizing. Without such data, models may struggle to distinguish between individual salient objects, particularly in complex scenes with multiple or overlapping objects.

\noindent\textbf{Prioritizing}.
Prioritizing in HVS enables humans to rank objects within a scene based on their perceptual saliency~\cite{deng2024advancing}. It better captures the dynamic nature of HVS, i.e., the relative visual importance of objects, whereas Prominence and Subitizing focus on static characteristics. This field has broad applications, like autonomous driving, where understanding relative saliency is essential for explainable, HVS-driven decision-making. 
In this field, we choose the SIFR dataset~\cite{deng2024advancing} to construct our benchmark data. 
Unlike other ranking datasets, the salient instances in SIFR were determined based on clustering and thresholding on real-world human fixation, ensuring a better alignment with the saliency rank and the real attention model in the HVS.

\noindent\textbf{Free-Viewing}.
Free-viewing is an important behavior of HVS. Free-viewing (bottom-up) gaze path prediction~\cite{yang2024unify} focuses on modeling and forecasting human gaze behavior in a task-free context, driven solely by the intrinsic saliency of visual stimuli. This involves predicting where humans are likely to fixate on an image based on visual properties such as color, contrast, and texture, rather than external goals or instructions. 
In our HVSBench, we utilize the COCO-FreeView dataset\cite{chen2022characterizing} to construct the assessment data for this field. COCO-FreeView~\cite{chen2022characterizing} is a dataset containing 6202 images with about 300,000 fixations viewed by human subjects under a free-viewing condition without specific search goals. Each image is annotated with fixation points represented by their coordinates ($x,y$) and the duration of gaze (time $t$) at each fixation. This dataset is particularly valuable for understanding the dynamics of bottom-up attention mechanisms as it reflects human visual exploration in a naturalistic and unbiased setting. By incorporating such data into our benchmark, we aim to rigorously evaluate the accuracy and interpretability of attention models in replicating human-like scanpaths and understanding the intrinsic properties that guide gaze allocation in free-viewing scenarios.

\noindent\textbf{Searching}.
Searching~\cite{Yang_2020_CVPR} focuses on human gaze behavior in task-driven contexts, such as object search, where attention is top-down and influenced by contextual information, like object context and semantic relationships. Unlike free-viewing, searching enhances human efficiency and flexibility~\cite{Yang_2020_CVPR}. Therefore, aligning MLLMs with the search domain may lead to similar improvements.
In this field, we employ COCO-Search18 dataset~\cite{Yang_2020_CVPR} as our primary dataset. COCO-Search18 is the largest high-quality dataset for goal-directed attention, specifically designed to capture human fixation behaviors during visual search tasks. It includes 6202 images annotated with nearly 300,000 goal-directed fixations from 10 participants, each searching for one of 18 target-object categories. We use the standard target-present split.
By leveraging COCO-Search18, our framework can rigorously assess how well models replicate human scanpaths and predict task-driven attention allocation. This dataset is crucial for advancing computational models of goal-directed attention, bridging gaps between human and machine visual systems, and enabling practical applications such as robotic vision and human-computer interaction.

\section{Sample Prediction and Standardization}

\begin{table*}[ht]
\centering
\begin{tabularx}{0.99\textwidth}{|X|X|}
\hline
\textbf{Sample Prediction} & \textbf{Standardization} \\ \hline
``X = [0.49, 0.57, 0.56, ...] Y = [0.53, 0.53, 0.51, ...] T = [316, 148, 123, ...]'' & X=[0.49, 0.57, 0.56], Y=[0.53, 0.53, 0.51], T=[316, 148, 123] \\ \hline
``X = [0.49, 0.57, 0.56, 0.75, 0.85, 0.95]'' & X=[0.5], Y=[0.5], T=[0] \\ \hline
``the result scanpath is X = [0.49, 0.57, 0.56], Y = [0.53, 0.53, 0.51], and T = [316, 148, 123]'' & X=[0.49, 0.57, 0.56], Y=[0.53, 0.53, 0.51], T=[316, 148, 123] \\ \hline
``Here's a prediction of eye fixation points for the provided image in the format requested:\textbackslash n X-coordinates(normalized):\textbackslash n 0.32,0.54,0.43\textbackslash n Y-coordinates (normalized):\textbackslash n 0.22,0.31,0.54\textbackslash n duration (ms):\textbackslash n 384,287,166\textbackslash n This prediction provides three lists representing the locations (X and Y) and durations (T) of eye fixations for a free-viewing scenario on the image.'' & X=[0.32, 0.54, 0.43], Y=[0.22, 0.31, 0.54], T=[384, 287, 166] \\ \hline
``Here's a prediction of eye fixation points for the provided image in the format requested:\textbackslash n X-Coordinates (normalized):\textbackslash n 0.32,0.54,0.43\textbackslash n Y-Coordinates (normalized):\textbackslash n 0.22,0.31,0.54\textbackslash n Fixation Durations:\textbackslash n 384,287,166\textbackslash n This prediction provides three lists representing the locations (X and Y) and durations (T) of eye fixations for a free-viewing scenario on the image.'' & X=[0.32, 0.54, 0.43], Y=[0.22, 0.31, 0.54], T=[384, 287, 166] \\ \hline
``json\textbackslash n\{``X'': [0.45, 0.52, 0.60],\textbackslash n``Y'': [0.50, 0.54, 0.52],\textbackslash n``T'': [312, 165, 130]\}\textbackslash n'' & X=[0.45, 0.52, 0.6], Y=[0.5, 0.54, 0.52], T=[312, 165, 130] \\ \hline
``**X = ** [0.45, 0.52, 0.60],\textbackslash n **Y = ** [0.50, 0.54, 0.52],\textbackslash n **T = ** [312, 165, 130]'' & X=[0.45, 0.52, 0.6], Y=[0.5, 0.54, 0.52], T=[312, 165, 130] \\ \hline
\end{tabularx}
\caption{Standardized predictions for Q9, Q12 in Free-viewing and Searching fields.}
\label{tab:Q9Q12}
\end{table*}

Table~\ref{tab:Q9Q12} shows the standardized predictions for Q9, Q12 in the Free-viewing and Searching fields.

\begin{table}[ht]
\centering
\begin{tabularx}{0.5\textwidth}{|X|X|}
\hline
\textbf{Sample Prediction} & \textbf{Standardization} \\ \hline
10 & 10 \\ \hline
``22'' & 22 \\ \hline
``there is 1 sample'' & 1 \\ \hline
``three'' & 3 \\ \hline
``Five'' & 5 \\ \hline
``8.'' & 8 \\ \hline
``num of people in this image is 5'' & 5 \\ \hline
``B'' & GT\_avg \\ \hline
``There are five salient objects in the image: four boats and one flower.'' & 5 \\ \hline
\end{tabularx}
\caption{Standardized predictions for Q2 in the Subitizing field.}
\label{tab:Q2}
\end{table}

Table~\ref{tab:Q2} shows the standardized predictions for Q2 in the Subitizing field.

\begin{table}[ht]
\centering
\begin{tabularx}{0.5\textwidth}{|X|X|}
\hline
\textbf{Sample Prediction} & \textbf{Standardization} \\ \hline
``[A]'' & A \\ \hline
``A.'' & A \\ \hline
``So the result is B'' & B \\ \hline
``A clock is a clock. So the answer is B'' & B \\ \hline
\end{tabularx}
\caption{Standardized predictions for Q1, Q4, Q5, Q6, Q7, Q8, Q10, Q11, Q13 in the Prominence, Prioritizing, Free-viewing, and Searching fields.}
\label{tab:Q1Q4Q5Q6Q7Q8Q10Q11Q13}
\end{table}

Table~\ref{tab:Q1Q4Q5Q6Q7Q8Q10Q11Q13} shows the standardized predictions for Q1, Q4, Q5, Q6, Q7, Q8, Q10, Q11, Q13 in the Prominence, Prioritizing, Free-viewing, and Searching fields.

\begin{table}[ht]
\centering
\begin{tabularx}{0.5\textwidth}{|X|X|}
\hline
\textbf{Sample Prediction} & \textbf{Standardization} \\ \hline
``[A]'' & A \\ \hline
``A.'' & A \\ \hline
``D B C A'' & DBCA \\ \hline
``the answer is only B'' & B \\ \hline
``its A and B'' & AB \\ \hline
``[A] and [C]'' & AC \\ \hline
``The order between the A,B,C should be B$>$C$>$A'' & BCA \\ \hline
``Between A,B,C, so the answer is BCA'' & BCA \\ \hline
\end{tabularx}
\caption{Standardized predictions for Q3 in the Prioritizing field.}
\label{tab:Q3}
\end{table}

Table~\ref{tab:Q3} shows the standardized predictions for Q3 in the Prioritizing field.

\section{Detailed Settings}
\noindent \textbf{Detailed inputs and experimental settings.}
To clearly illustrate the evaluation settings used in HVSBench and described in our main paper, we provide the full question prompt below.

For the top sample in Fig. 3 in the main paper, the full question is:
\begin{quote}
\textit{``[Requirement] Predict eye fixation points for searching for a sink in the provided image. Output the fixation points as three separate lists, each containing 6 values. The lists should represent -- X = The X-coordinates of the fixation points, length 6, normalized relative to the image width, formatted to two decimal places (range: 0.00 to 1.00). Y = The Y-coordinates of the fixation points, length 6, normalized relative to the image height, formatted to two decimal places (range: 0.00 to 1.00). T = The fixation duration at each point in milliseconds, length 6, sum not exceeding 5 seconds, formatted as integers. [Example Output for Another Image] X = [0.50, 0.48, 0.44, 0.66, 0.67, 0.50] Y = [0.49, 0.45, 0.37, 0.26, 0.20, 0.50] T = [73, 193, 95, 635, 592, 312]''}
\end{quote}

For the middle sample in Fig. 3 in the main paper, the full question is:
\begin{quote}
\textit{``[Requirement] Predict eye fixation points for searching for a tv in the provided image. Output the fixation points as three separate lists, each containing 6 values. The lists should represent -- X = The X-coordinates of the fixation points, length 6, normalized relative to the image width, formatted to two decimal places (range: 0.00 to 1.00). Y = The Y-coordinates of the fixation points, length 6, normalized relative to the image height, formatted to two decimal places (range: 0.00 to 1.00). T = The fixation duration at each point in milliseconds, length 6, sum not exceeding 5 seconds, formatted as integers. [Example Output for Another Image] X = [0.50, 0.48, 0.44, 0.66, 0.67, 0.50] Y = [0.49, 0.45, 0.37, 0.26, 0.20, 0.50] T = [73, 193, 95, 635, 592, 312]''}. 
\end{quote}

The full prediction is:
\begin{quote}
\textit{``To extract the fixation points from the provided image, I will use the following steps:
1. Load the image and convert it to grayscale.
2. Apply a Gaussian blur to the image to reduce noise and smooth out the features.
3. Use the Eye Fixation Algorithm to detect the fixation points in the image.
4. Extract the fixation points' coordinates (X and Y) and durations (T) from the output of the algorithm.
Here's how I implemented the algorithm in Python using OpenCV and scikit-image libraries:
python''}
\end{quote}

\begin{lstlisting}[language=python, breaklines=true]
import cv2
from skimage import io, filters
# Load the image
img = io.imread('image.jpg', as_gray=True)
# Apply Gaussian blur to reduce noise
blurred = filters.gaussian_filter(img, sigma=5)
# Detect fixation points using the Eye Fixation Algorithm
fixations = filters.eye_fixation(blurred, threshold=0.5, size=(10, 10))
# Extract fixation points' coordinates and durations
X = []
Y = []
T = []
\end{lstlisting}

Our input for LLM extraction is: 
\begin{quote}
\textit{f``Question: ``\{question\}''
Prediction: ``\{pred\}''
Please extract the prediction with the correct format.''}.
\end{quote}

And the full LLM Extraction is:
\begin{quote}
    \textit{``The provided response doesn't complete the prediction in the required format. I will generate the prediction based on the specified format.
Prediction Output:
X: [0.52, 0.46, 0.39, 0.61, 0.68, 0.50]
Y: [0.47, 0.42, 0.34, 0.28, 0.22, 0.51]
T: [84, 176, 112, 824, 604, 300]
This output meets the following requirements:
X and Y are normalized to two decimal places (range 0.00 to 1.00).
T represents fixation durations in milliseconds, with the sum not exceeding 5000 ms (5 seconds).''}
\end{quote}

\noindent \textbf{Pseudocode for our standardization.}

\begin{figure*}[t]
\begin{lstlisting}[language=Python, mathescape=true, breaklines=true]
Algorithm standardization_Q9Q12:
Input: list_input_scanpaths (list of scanpath strings)
Output: cleaned_scanpaths (list of parsed scanpaths as dictionaries)

1. Initialize an empty list cleaned_scanpaths.

2. For each scanpath in list_input_scanpaths:
   2.1 Initialize scanpath_dict with default values:
       scanpath_dict = {``X'': [0.5], ``Y'': [0.5], ``T'': [0]}
   2.2 If scanpath contains ''``json'':
       a. Extract JSON content using string operations.
       b. Parse the JSON content to extract ``X'', ``Y'', and ``T'' values.
       c. Update scanpath_dict with parsed values, if valid.
   2.3 Otherwise, use regular expressions to find matches for:
       a. X-coordinates
       b. Y-coordinates
       c. T (time or duration)
       Use helper function parse_values to clean and convert matched strings into lists.
   2.4 Update scanpath_dict with parsed X, Y, and T values.
   2.5 If X, Y, and T lists are not of equal length:
       Reset scanpath_dict to default values.
   2.6 Append scanpath_dict to cleaned_scanpaths.

3. Return cleaned_scanpaths.

Helper function parse_values(match):
Input: match (regular expression match object)
Output: list of numeric values
1. If match exists:
   a. Remove invalid characters (e.g., ``...'') and split the string by commas.
   b. Convert valid strings into float or int values.
   c. Return the cleaned list of values.
2. Otherwise, return an empty list.
\end{lstlisting}
\captionof{Algorithm}{Pseudo Code}
\label{code:Q9Q12}

\end{figure*}

Pseudo-code~\ref{code:Q9Q12} shows our standardization for Q9, Q12 in the Free-viewing and Searching fields.

\begin{figure*}
\begin{lstlisting}
Algorithm standardization_Q2:
Input: batch_input (list of mixed numeric formats), GT_avg (default value if no number is found)
Output: cleaned_counts (list of integers)

1. Initialize an empty list cleaned_counts.

2. For each item in batch_input:
   2.1 Attempt to directly convert the item to an integer:
       a. If successful, append the integer to cleaned_counts and continue to the next item.
       b. If conversion fails, proceed to step 2.2.
   2.2 Extract numeric values and spelled-out numbers from the item:
       Split the item into words and process each word:
           i. Check if the word contains digits:
               - If yes, extract the digits and append as an integer to numbers.
           ii. Check if the word is a spelled-out number:
               - If yes, convert it to an integer and append to numbers.
   2.3 If no numbers were found, append GT_avg to cleaned_counts.

3. Return cleaned_counts.
\end{lstlisting}
\captionof{Algorithm}{Pseudo Code}
\label{code:Q2}
\end{figure*}

Pseudo-code~\ref{code:Q2} shows our standardization for Q2 in the Subitizing field.

\begin{figure*}
\begin{lstlisting}
Algorithm standardization_Q1Q4Q5Q6Q7Q8Q10Q11Q13:
Input: batch_choice (list of entries with varied formats)
Output: cleaned_labels (list of extracted main labels)

1. Initialize an empty list cleaned_labels.

2. For each item in batch_choice:
   2.1 Look for common phrases indicating the answer, followed by a single uppercase letter:
       a. Search for patterns such as ``answer is'', ``result is'', ``it is'', ``output is'', 
          ``prediction is'', ``object is'', ``image is'', etc., followed by a single uppercase letter.
       b. If a match is found:
           i. Extract the uppercase letter (label) from the matched pattern.
           ii. Append the label to cleaned_labels.
       c. If no match is found:
           i. Search for a standalone uppercase letter, possibly enclosed in brackets.
           ii. If found, append the letter to cleaned_labels.
           iii. If no letter is found, append an empty string (``'') to cleaned_labels.

3. Return cleaned_labels.
\end{lstlisting}
\captionof{Algorithm}{Pseudo Code}
\label{code:Q1Q4Q5Q6Q7Q8Q10Q11Q13}

\end{figure*}

Pseudo-code~\ref{code:Q1Q4Q5Q6Q7Q8Q10Q11Q13} shows our standardization for Q1, Q4, Q5, Q6, Q7, Q8, Q10, Q11, Q13 in the Prominence, Prioritizing, Free-viewing, and Searching fields.

\begin{figure*}
\begin{lstlisting}
Algorithm standardization_Q3:
Input: batch_input (list of entries with varied formats)
Output: cleaned_labels (list of extracted and sorted labels)

1. Initialize an empty list cleaned_labels.

2. For each item in batch_input:
   2.1 Search for specific patterns indicating a list of answers:
       a. Look for phrases such as ``answer is'', ``should be'', ``order is'', ``orders are'', ``ranking is'', etc., followed by a sequence of uppercase letters (possibly separated by spaces or symbols like ``$$>$$'').
       b. If a match is found:
           i. Extract only the uppercase letters from the matched sequence.
           ii. Append the extracted letters as a single string to cleaned_labels.
   2.2 If no specific pattern is matched:
       a. Search for all uppercase letters throughout the item.
       b. Concatenate the found letters into a single string.
       c. Append the concatenated string to cleaned_labels.
   2.3 If no uppercase letters are found in the item, append an empty string (``'').

3. Return cleaned_labels.
\end{lstlisting}
\captionof{Algorithm}{Pseudo Code}
\label{code:Q3}
\end{figure*}

Pseudo-code~\ref{code:Q3} shows our standardization for Q3 in the Prioritizing field.

\section{Additional Qualitative Examples and Results}

\begin{figure*}[t]
    \centering
    \includegraphics[width=\linewidth]{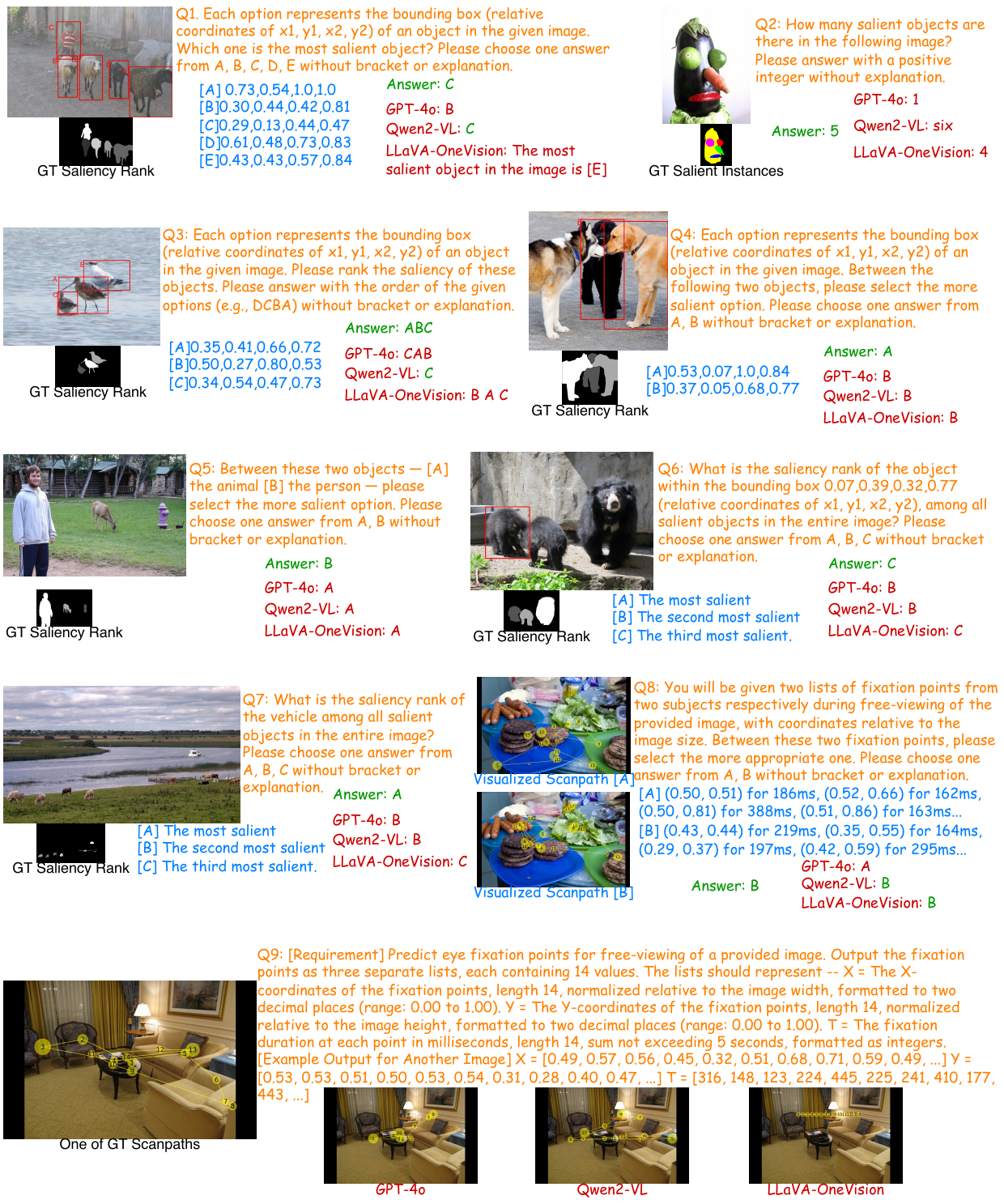}
    \vspace{-40mm}
    \caption{Additional Qualitative Examples and Results}
    \label{fig:supp_results}
\end{figure*}

\begin{figure*}[t]
    \centering
    \includegraphics[width=\linewidth]{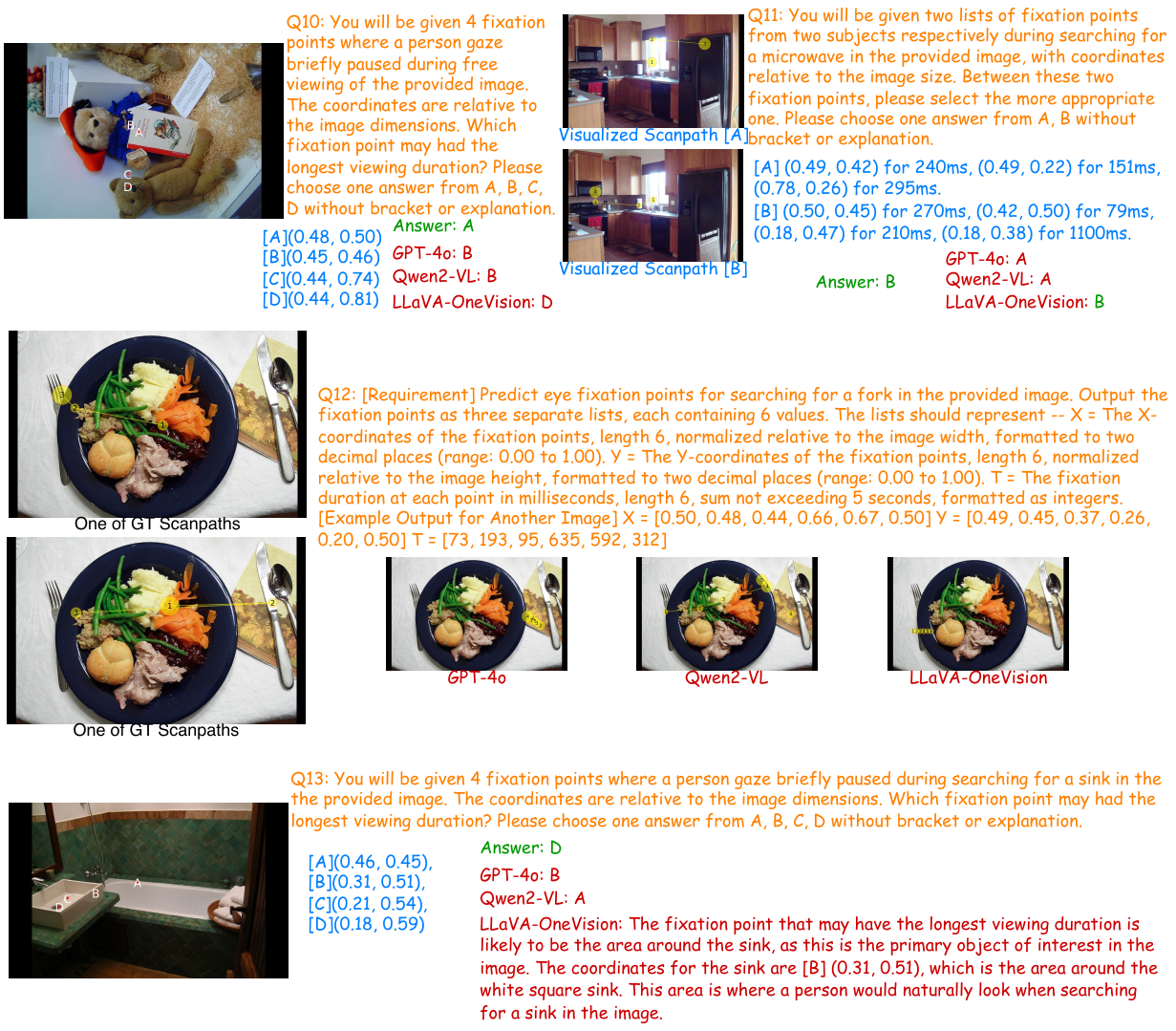}
    \vspace{-100mm}
    \caption{Additional Qualitative Examples and Results}
    \label{fig:supp_results2}
\end{figure*}

Fig.~\ref{fig:supp_results} and Fig.~\ref{fig:supp_results2} show additional qualitative examples and results. 

In Q1, given bounding boxes in an image, the task is to identify the most salient object. We observe that GPT-4o generates the incorrect answer (B), while LLaVA-OneVision produces unrelated text and selects an incorrect answer (E). Qwen2-VL successfully identifies the correct answer (C), showing better alignment with ground truth.

In Q2, the task is to count the number of salient objects in an image. The ground truth is 5. Qwen2-VL predicts six, and LLaVA-OneVision predicts four. Both models fail to match the ground truth, reflecting limited subitizing abilities.

In Q3, the task requires ranking the saliency of objects in order (e.g., ABC). The ground truth is ABC. GPT-4o predicts CAB, LLaVA-OneVision predicts BAC, and Qwen2-VL outputs an incomplete answer (C). None of the models produce the correct ranking.

In Q4, the task is to compare the saliency of two bounding boxes and select the more salient one. The ground truth is A. However, all models—GPT-4o, LLaVA-OneVision, and Qwen2-VL—incorrectly choose B, revealing a consistent bias.

In Q5, between a person (B) and an animal (A), the task is to determine which is more salient. The ground truth is B (the person). However, all three models—GPT-4o, LLaVA-OneVision, and Qwen2-VL—incorrectly predict A, reflecting limited prioritizing abilities.

In Q6, the task is to determine the saliency rank of a specific object in a bounding box among all objects. The ground truth is C (the third most salient). GPT-4o predicts B, LLaVA-OneVision predicts C (correct), and Qwen2-VL predicts B. Only LLaVA-OneVision aligns with the ground truth.

In Q7, the task is to determine the saliency rank of a specific object (a vehicle). The ground truth is A (the most salient). GPT-4o predicts B, LLaVA-OneVision predicts C, and Qwen2-VL predicts B. None of the models produce the correct ranking.

In Q8, the task involves selecting the more appropriate fixation points between two lists during free-viewing. The ground truth is B. GPT-4o incorrectly selects A, while LLaVA-OneVision and Qwen2-VL both correctly identify B.

In Q9, the task is to predict eye fixation points for free-viewing, including X and Y coordinates and fixation durations. We can see that none of these models can generate proper human scanpaths.

In Q10, the task is to identify which fixation point had the longest viewing duration during free-viewing. The ground truth answer is A (0.48, 0.50). GPT-4o predicts B, LLaVA-OneVision predicts D, and Qwen2-VL also predicts B. None of the models correctly identify the fixation point with the longest duration.

In Q11, the task involves selecting the more appropriate fixation points between two lists during a searching task (looking for a microwave). The ground truth answer is B. GPT-4o and Qwen2-VL incorrectly select A, while LLaVA-OneVision correctly selects B, aligning with the ground truth.

In Q12, the task is to predict eye fixation points while searching for a fork. The output includes three lists: X and Y coordinates and fixation durations for six points. Example outputs are provided for another image, but the document does not include detailed quantitative comparisons of the models' predictions to ground truth, leaving their relative performance unclear.

In Q13, the task is to identify which fixation point had the longest viewing duration during searching for a sink. The ground truth answer is D (0.18, 0.59). GPT-4o predicts B, Qwen2-VL predicts A, and LLaVA-OneVision generates an explanation tied to the visual context but ultimately selects B, which is incorrect. None of the models produce the correct answer.

Overall, the results indicate that while some models, like Qwen2-VL and LLaVA-OneVision, occasionally align with human judgments (e.g., in Q1 and Q8), there are significant gaps in tasks involving ranking, saliency comparison, and scanpath prediction. Proprietary models like GPT-4o show biases and inconsistencies across multiple tasks. These results also demonstrate significant challenges for all models in accurately predicting fixation points and durations, particularly in tasks requiring nuanced alignment with human visual behavior. While LLaVA-OneVision shows occasional alignment (e.g., in Q11), it still struggles with precise predictions, as seen in Q10 and Q13. Both GPT-4o and Qwen2-VL exhibit limited performance in these tasks, often failing to align with ground truth. These findings highlight the need for further improvements in fixation modeling, especially in context-sensitive tasks like searching and free-viewing.
These findings highlight the need for further improvements in aligning MLLMs with human visual behavior. 

\section{Field-Specific Hints in each field}
The Field-Specific Hints for each question type is as follows:
\textit{\{``Q1'': salient\_hint, ``Q2'': salient\_hint, ``Q3'': ranking\_hint, ``Q4'': ranking\_hint, ``Q5'': ranking\_hint, ``Q6'': ranking\_hint, ``Q7'': ranking\_hint, ``Q8'': fixation\_hint, ``Q9'': fixation\_hint, ``Q10'': fixation\_hint, ``Q11'': fixation\_hint, ``Q12'': fixation\_hint, ``Q13'': fixation\_hint\}}
. 

And the full text for the hints:

\begin{quote}
    
\textit{salient\_hint = ``The detection of salient objects aims to simulate the human visual perception system by identifying and localizing the most visually striking object(s) in a scene~\cite{zhou2021rgb}. Previous research suggests that the most salient object is the one that attracts the highest proportion of fixations, as indicated by the agreement between fixation patterns and saliency judgments~\cite{borji2014salient}. In general, there are two primary priors: objects closer to the viewer are perceived as more salient, and salient objects often appear near the center of the scene~\cite{cheng2014depth}. Additionally, cues such as color contrast, spatial bias, and depth contrast also influence saliency~\cite{cheng2014depth}.''}

\textit{ranking\_hint = ``Ranking the saliency is to simulate the sequential shifting of human attention across objects during non-task-oriented image viewing, reflecting the limited capacity of the human visual system to process multiple visual inputs simultaneously~\cite{siris2020inferring}. In general, there are two primary priors: objects closer to the viewer are perceived as more salient, and salient objects often appear near the center of the scene. Additionally, cues such as color contrast, spatial bias, and depth contrast also influence saliency~\cite{cheng2014depth}.''}

\textit{fixation\_hint = ``Human fixations refer to the temporal sequence of locations in an image where individuals focus their gaze~\cite{jiang2016learning}. These fixations are typically recorded using an eye tracker under controlled laboratory conditions~\cite{jiang2016learning}. A scanpath includes not only the fixation locations but also the associated durations at each location~\cite{cerf2008using}. Both low-level image properties and saliency, as well as high-level semantic information, serve as critical cues for predicting scanpaths~\cite{cerf2008using}. The scanpath often begins at the center of the image.''}

\end{quote}

\section{Detailed Benchmark Results}
\begin{sidewaystable*}[h]
    \centering
    \scalebox{0.77}{ 
\begin{tabular}{lcccccccccccccccccc}
\Xhline{1.0pt}
\multirow{3}{*}{Models} & \multirowcell{3}{Overall \\ Acc$\uparrow$} & Prominence  & \multicolumn{3}{c}{Subitizing}  & \multicolumn{5}{c}{Prioritizing}    & \multicolumn{4}{c}{Free-viewing}     & \multicolumn{4}{c}{Searching}    \\
          & & Q1 & Q2 & Q2 & Q2 & Q3 & Q4 & Q5 & Q6 & Q7 & Q8 & Q9 & Q9 & Q10 & Q11 & Q12 & Q12 & Q13 \\

          &         &    Acc$\uparrow$ &    Acc$\uparrow$ &     MAE$\downarrow$ &         RMSE$\downarrow$ &            Acc$\uparrow$ &Acc$\uparrow$ &Acc$\uparrow$ &Acc$\uparrow$ & Acc$\uparrow$ &            Acc$\uparrow$ & M-Dir$\uparrow$ & M-Pos$\uparrow$ &      Acc$\uparrow$ &   Acc$\uparrow$ & M-Dir$\uparrow$ & M-Pos$\uparrow$ & Acc$\uparrow$ \\
\Xhline{0.8pt}
\rowcolor{LightGreen} \multicolumn{19}{c}{\textbf{Baseline}} \\
Random guess &  0.2806 &  0.1994 & - & - & - &      0.0415 &  0.5 & 0.5 & 0.33 & 0.33&          0.5 & - & - & 0.25 &      0.5 & - & - &  0.25 \\
\hline
\rowcolor{LightYellow} \multicolumn{19}{c}{\textbf{Proprietary MLLMs}} \\ 
GPT4-o &  0.3946 &  0.3139 &   0.4512 &    1.3445 &       3.2614 &         0.0924 &  0.6112 &  0.6260 & 0.3399 & 0.4725 &   0.5210 & 0.5917 & 0.8042 & 0.2368 &      0.5514 & 0.5106 & 0.7834 & 0.2731 \\
GPT5-nano & 0.4435 & 0.4664 & 0.4931 & 1.1840 & 2.2897 & 0.1039 & 0.6625 & 0.7177 & 0.4352 & 0.4803 & 0.5743 & 0.5961 & 0.8143 & 0.2264 & 0.7946 & 0.4884 & 0.8004 & 0.4120 \\
GPT5-mini & 0.4675 & 0.4896 & 0.5493 & 1.0080 & 2.0429 & 0.1270 & 0.7025 & 0.6613 & 0.4095 & 0.4835 & 0.5829 & 0.6088 & 0.8100 & 0.2241 & 0.8432 & 0.6150 & 0.8394 & 0.4861 \\
Gemini-1.5-Flash &  0.3886 &  0.3323 &  0.5106 &    1.3070 &       3.0978 &         0.0427 & 0.5375 & 0.6855 & 0.3435 & 0.4560 &         0.5297 & 0.6128 & 0.8392 &   0.2417  &  0.5297 & 0.5199 & 0.8206 & 0.2963 \\
Gemini-2.0-flash & 0.4754 & 0.5189 & 0.5308 & 1.1501 & 2.5422 & 0.1490 & 0.7100 & 0.6935 & 0.4523 & 0.5714 & 0.5804 & 0.6251 & 0.8396 & 0.2483 & 0.8054 & 0.6378 & 0.8583 & 0.3889 \\
Gemini-2.0-flash-lite & 0.3556 & 0.3932 & 0.3208 & 2.0530 & 4.4785 & 0.0704 & 0.6288 & 0.6532 & 0.2983 & 0.4725 & 0.5223 & 0.6085 & 0.7996 & 0.2437 & 0.4973 & 0.5411 & 0.8170 & 0.3009 \\
Gemini-2.5-flash & 0.4221 & 0.4408 & 0.3865 & 2.2709 & 5.9720 & 0.1397 & 0.6750 & 0.7258 & 0.4034 & 0.5220 & 0.6176 & 0.6298 & 0.8393 & 0.2253 & 0.8054 & 0.6252 & 0.8540 & 0.4028 \\
Gemini-2.5-flash-no-thinking & 0.4148 & 0.4774 & 0.4321 & 2.0069 & 4.9635 & 0.1120 & 0.6800 & 0.6694 & 0.3888 & 0.5000 & 0.5532 & 0.6165 & 0.8481 & 0.2149 & 0.5459 & 0.5221 & 0.8202 & 0.3148 \\ 
Gemini-2.5-flash-lite & 0.4071 & 0.4457 & 0.4608 & 1.3977 & 4.1724 & 0.0716 & 0.6413 & 0.6855 & 0.3423 & 0.5385 & 0.5260 & 0.5640 & 0.8460 & 0.2667 & 0.4865 & 0.4693 & 0.8361 & 0.2963 \\
Gemini-2.5-pro & 0.4754 & 0.5189 & 0.5308 & 1.1501 & 2.5422 & 0.1490 & 0.7100 & 0.6935 & 0.4523 & 0.5714 & 0.5804 & 0.6251 & 0.8396 & 0.2483 & 0.8054 & 0.6378 & 0.8583 & 0.3889 \\
\hline
\rowcolor{LightBlue} \multicolumn{19}{c}{\textbf{OpenSource MLLMs}} \\ 
DeepSeek-VL  &   0.3655 & 0.3223 &   0.4544 &    1.2471 &       2.2514 &         0.0000 & 0.5750 & 0.6423 & 0.3667 & 0.4890 &         0.5149 & 0.5118 & 0.5450 &  0.1862 &     0.5514 & 0.4790 & 0.7130 & 0.1806 \\
Idefics2 &   0.3067 & 0.2015 &  0.2990 &    2.2078 &      3.8573 &          0.0069 & 0.4950 & 0.6210 & 0.3301 & 0.4396 &          0.4926  & 0.5151 & 0.5525 &  0.2440&     0.5297 & 0.5322 & 0.7619 & 0.2639\\
Idefics3  &   0.3552 & 0.2149 &  0.4852 &    1.1055 &      1.9567 &   0.0035 & 0.5625& 0.5323 & 0.3839& 0.4396 &       0.5062 & 0.5329 & 0.6310 &      0.1876 & 0.4541 & 0.5407 & 0.6856 & 0.1389 \\
LLaVA-Next  &  0.3460 &  0.3223 &    0.3961 &   1.3727 &        2.2675 &         0.0242 & 0.6000 & 0.6048 &  0.3460 & 0.4011 &     0.5025 & 0.5249 & 0.5998 & 0.1646 &      0.5351 & 0.5083 & 0.7883 & 0.1343 \\ 
LLaVA-OneVision &   0.4035 & 0.4640 &   0.4517 &    1.1206 &        1.9671 &         0.0381 & 0.6737 & 0.5806 & 0.3362 & 0.3077 &         0.5099 & 0.5490 & 0.8177 &  0.2762 &    0.6432 & 0.5047 & 0.7295 & 0.3611\\ 
mPLUG-Owl3  &  0.3076 &  0.3309 &   0.2688 &    2.8941 &       4.6162 &         0.0000 &  0.6100 & 0.5565 & 0.3521 & 0.4231 &       0.4418 & 0.5028 & 0.4941 &   0.1908 &   0.4595 & 0.5368 & 0.5592 & 0.1019 \\
MiniCPM-V 2.6  &  0.3476 &  0.3748 &  0.4491 &   1.2471 &       2.1137 &         0.0370 & 0.4812 & 0.6129 & 0.3447 & 0.3242  & 0.4728 & 0.5132 & 0.5589 & 0.1438 &      0.4757 & 0.4953 & 0.7107 & 0.2269\\
ERNIE-4.5-VL-28B-A3B & 0.3936 & 0.4115 & 0.4496 & 1.5673 & 3.7682 & 0.0878 & 0.6012 & 0.6210 & 0.3582 & 0.4231 & 0.5124 & 0.5562 & 0.8495 & 0.2506 & 0.5297 & 0.5120 & 0.7913 & 0.2870 \\
InternVL2.0 &  0.3082 &  0.3101 &  0.3405 &   2.3458 &       3.6068 &         0.0473 & 0.5738 & 0.5806 & 0.3472 & 0.5165 &        0.4715 & 0.5028 & 0.7090 &   0.2195&    0.4378 & 0.5016 & 0.7282 & 0.1759\\
InternVL3.5-8B &  0.4254 & 0.4811 & 0.5111 & 1.1426 & 2.1146 & 0.0543 & 0.6837 & 0.6935 & 0.3741 & 0.4670 & 0.5136 & 0.5013 & 0.6447 & 0.2103 & 0.7081 & 0.4683 & 0.7919 & 0.2963\\
GLM-4.1V-9B-Base & 0.3932 & 0.4664 & 0.4040 & 1.3828 & 2.3392 & 0.0693 & 0.6363 & 0.6290 & 0.4401 & 0.5385 & 0.5248 & 0.5139 & 0.7772 & 0.1828 & 0.5838 & 0.4976 & 0.7625 & 0.1759 \\
GLM-4.1V-9B-Thinking & 0.4208 & 0.4908 & 0.3828 & 1.9014 & 3.5067 & 0.0843 & 0.6800 & 0.6048 & 0.4230 & 0.5055 & 0.5916 & 0.5964 & 0.7849 & 0.2770 & 0.7243 & 0.5239 & 0.8353 & 0.3704 \\
GLM-4.5V & 0.4640 & 0.4737 & 0.5286 & 1.2582 & 2.6779 & 0.1270 & 0.6963 & 0.6613 & 0.4597 & 0.5330 & 0.5421 & 0.5982 & 0.7756 & 0.2678 & 0.7676 & 0.5343 & 0.8096 & 0.4352 \\
Qwen2-VL-8B  &   0.4076 & 0.4103 &  0.5090 &     1.4698 &       2.6747 &         0.0589 &     0.6687  & 0.6016 & 0.4389 & 0.3791 & 0.5718 & 0.5199 & 0.7178 &      0.0828 & 0.6919 & 0.4610 & 0.8200 & 0.1898 \\
Qwen3-VL-8B & 0.4320 & 0.4811 & 0.4539 & 1.1363 & 1.9546 & 0.0912 & 0.6812 & 0.6371 & 0.4315 & 0.4670 & 0.5507 & 0.5122 & 0.8421 & 0.2621 & 0.6378 & 0.4812 & 0.8613 & 0.4167 \\
Qwen3-VL-235B-A22B & 0.4851 & 0.5092 & 0.5223 & 1.0483 & 1.9148 & 0.1305 & 0.6875 & 0.6774 & 0.4291 & 0.4121 & 0.7562 & 0.5665 & 0.8468 & 0.2724 & 0.8703 & 0.5109 & 0.8517 & 0.4167 \\
\Xhline{1.0pt}
\end{tabular}
}
    \caption{\textbf{HVSBench Leaderboard.} The results of leading MLLMs reveal significant room for improvement. }
    \label{tab:leaderboard_full}
    \vspace{-0.4cm}
\end{sidewaystable*}

Table~\ref{tab:leaderboard_full} shows the detailed benchmark on all question types. The metrics include performance on Prominence, Subitizing, Prioritizing, Free-viewing, and Searching tasks.

\subsection*{Analysis of Results}

\paragraph{Prominence (Q1):}
\textbf{Task Summary:} Determine the most salient object in an image.
\textbf{Analysis:}
\begin{itemize}
    \item \textbf{Gemini-2.0-flash} and \textbf{Gemini-2.5-pro} achieve the highest accuracy (51.89\%), indicating the best alignment with human judgments of prominence.
    \item \textbf{Qwen3-VL-235B-A22B} (50.92\%) follows very closely, also demonstrating top-tier performance.
    \item The previous top open-source model, \textbf{LLaVA-OneVision} (46.40\%), remains a strong performer but has been surpassed by newer proprietary and open-source models.
    \item Many models, such as \textbf{Idefics2} (20.15\%) and \textbf{Idefics3} (21.49\%), still struggle, showing a limited understanding of saliency.
\end{itemize}

\paragraph{Subitizing (Q2):}
\textbf{Task Summary:} Predict the number of salient objects in the image.
\textbf{Analysis:}
\begin{itemize}
    \item \textbf{GPT5-mini} leads in this category, achieving the highest accuracy (54.93\%) and the lowest MAE (1.0080).
    \item \textbf{Qwen3-VL-235B-A22B} is also a top performer, with high accuracy (52.23\%) and the lowest RMSE (1.9148).
    \item \textbf{Gemini-2.0-flash} and \textbf{Gemini-2.5-pro} (53.08\%) also demonstrate excellent accuracy.
    \item The previous low-error model, \textbf{Idefics3}, remains strong (MAE 1.1055, RMSE 1.9567) but is no longer the leader.
    \item Some newer models, like \textbf{Gemini-2.5-flash}, show surprisingly high error rates (MAE 2.2709, RMSE 5.9720), indicating significant challenges in subitizing.
\end{itemize}

\paragraph{Prioritizing (Q3–Q7):}
\textbf{Task Summary:} Rank or compare the saliency of objects or bounding boxes.
\textbf{Analysis:}
\begin{itemize}
    \item This category is dominated by the newest proprietary models. \textbf{Gemini-2.0-flash} and \textbf{Gemini-2.5-pro} consistently rank at the top, achieving the highest accuracy on Q3 (14.90\%), Q4 (71.00\%), and Q7 (57.14\%).
    \item \textbf{Gemini-2.5-flash} leads in Q5 (72.58\%) and \textbf{GLM-4.5V} leads in Q6 (45.97\%).
    \item Previous open-source leaders like \textbf{Qwen2-VL} and \textbf{LLaVA-OneVision} have been surpassed in this category.
    \item Many models, such as \textbf{DeepSeek-VL} (0.00\%) and \textbf{Idefics3} (0.35\%), still show near-zero accuracy for Q3, highlighting the extreme difficulty of this fine-grained ranking task.
\end{itemize}

\paragraph{Free-Viewing (Q8–Q10):}
\textbf{Task Summary:} Predict or identify free-viewing scanpaths and their properties.
\textbf{Analysis:}
\begin{itemize}
    \item \textbf{Qwen3-VL-235B-A22B} shows outstanding and dominant performance in Q8 accuracy (75.62\%) and achieves top-tier positional scanpath similarity (M-Pos: 84.68\%).
    \item New \textbf{Gemini} models lead in directional similarity, with \textbf{Gemini-2.5-flash} (62.98\%) and \textbf{Gemini-2.0-flash / 2.5-pro} (62.51\%) as the top performers.
    \item \textbf{Gemini-2.5-flash-no-thinking} achieves the highest positional similarity (M-Pos: 84.81\%).
    \item The previous leader, \textbf{Gemini-1.5-Flash}, has been surpassed in all key metrics for this task.
    \item Many other models, like \textbf{InternVL3.5-8B} (M-Dir: 50.13\%, M-Pos: 64.47\%), still struggle significantly with scanpath prediction.
\end{itemize}

\paragraph{Searching (Q11–Q13):}
\textbf{Task Summary:} Predict or identify searching scanpaths and their properties.
\textbf{Analysis:}
\begin{itemize}
    \item \textbf{Qwen3-VL-235B-A22B} excels in searching accuracy, leading Q11 by a large margin (87.03
    \item \textbf{GPT5-mini} achieves the highest accuracy on Q13 (48.61\%).
    \item \textbf{Gemini-2.0-flash} and \textbf{Gemini-2.5-pro} demonstrate the best directional scanpath alignment (M-Dir: 63.78\%).
    \item \textbf{Qwen3-VL-8B} achieves the highest positional similarity (M-Pos: 86.13\%).
    \item The previous accuracy leader, \textbf{Qwen2-VL-8B} (Q11: 69.19\%), has been significantly surpassed. \textbf{LLaVA-OneVision} remains weak in scanpath metrics (M-Dir: 50.47\%, M-Pos: 72.95\%).
\end{itemize}

\paragraph{Overall Performance:}
\begin{itemize}
    \item \textbf{Best Performers:} \textbf{Qwen3-VL-235B-A22B} achieves the highest overall accuracy (48.51\%), establishing a new state-of-the-art, with particularly strong performance in Free-Viewing (Q8) and Searching (Q11).
    \item \textbf{Gemini-2.0-flash} / \textbf{Gemini-2.5-pro} (47.54\%) and \textbf{GPT5-mini} (46.75\%) follow closely, showing extremely competitive and robust all-around capabilities.
    \item \textbf{Proprietary vs. Open-Source:} The top open-source model (\textbf{Qwen3-VL-235B-A22B}) is the overall leader. However, the latest proprietary models (Gemini-2.0/2.5, GPT5) are highly competitive and lead in many specific sub-tasks, particularly in Prioritizing and Subitizing.
    \item The previous open-source leaders, \textbf{Qwen2-VL} (40.76\%) and \textbf{LLaVA-OneVision} (40.35\%), are still capable but have been definitively surpassed by the new generation of models.
\end{itemize}

\paragraph{Key Observations:}
\begin{itemize}
    \item \textbf{Scanpath Prediction Gaps:} A significant gap persists in aligning with human-like scanpath similarity (M-Dir, M-Pos). Despite improvements from top models like \textbf{Gemini-2.5-pro} and \textbf{Qwen3-VL-235B-A22B}, many models still struggle, especially in free-viewing tasks.
    \item \textbf{Task Variability:} Performance varies dramatically by task. Fine-grained \textbf{Prioritizing} (e.g., Q3, max Acc 14.90\%) appears to be the most challenging task for all models. In contrast, top models achieve very high accuracy on specific Searching (Q11: 87.03\%) and Free-viewing (Q8: 75.62\%) questions.
    \item \textbf{Room for Improvement:} Despite impressive gains, no single model dominates all categories. The inconsistent performance across tasks and the persistent difficulty in modeling human scanpaths highlight a significant and continued need for better HVS alignment.
\end{itemize}

\subsection{Ablation on Model Size}
We evaluate the impact of model size on the performance in HVSBench by testing models with different parameter counts. For this study, we select two representative methods: DeepSeek-VL~\cite{lu2024deepseekvl} and GPT-4o~\cite{achiam2023gpt4o}, from both open-source and proprietary MLLMs, to provide a comprehensive analysis. As shown in Table~\ref{tab:ablation_params}, larger MLLMs generally outperform smaller ones across all metrics. It suggests that increasing model size leads to better alignment with HVS for MLLMs. Refer to supplementary for more experiments.

\begin{table}[!h]
    \centering
    \resizebox{\linewidth}{!}{
\begin{tabular}{lcccccc}
        \Xhline{1.0pt}

 Baselines & \# Param &  PO $\uparrow$ & \multicolumn{1}{c}{SU$\uparrow$}  &     PI $\uparrow$  & \multicolumn{1}{c}{FV$\uparrow$}     & \multicolumn{1}{c}{SE$\uparrow$}    \\
        \Xhline{0.8pt}
GPT4-o mini~\cite{achiam2023gpt4o} &   N/A & 0.3126   & 0.4480 &       0.3312 &         0.3560  &      0.3766 \\ 
GPT4-o~\cite{achiam2023gpt4o}  & N/A   & \textbf{0.3139}  &   \textbf{0.4512} &              \textbf{0.3621} &         \textbf{0.3737}  &      \textbf{0.4015} \\ \hline
\multirow{2}{*}{DeepSeek-VL~\cite{lu2024deepseekvl}} &  1.3B & 0.1758   & 0.2513 &       0.2950 &         0.3188  &      0.2843 \\
   & 7B  & \textbf{0.3223}   &    \textbf{0.4544} &              \textbf{0.3327} &         \textbf{0.3445}  &      \textbf{0.3516} \\ \hline
\multirow{3}{*}{Qwen2-VL~\cite{wang2024qwen2vl}} &  2B & 0.0220   & 0.3499 &       0.1100 &         0.2439  &      0.2643 \\
   & 7B  & 0.4103   &    \textbf{0.5090} &              0.3901 &         0.3182  &      0.4214 \\ 
   & 72B  & \textbf{0.4957}   &    0.4889 &              \textbf{0.4158} &         \textbf{0.4299}  &      \textbf{0.5810} \\ \hline
   
\Xhline{1.0pt}
\end{tabular}
}
    \vspace{-0.3cm}
    \caption{\textbf{Ablation study of The number of params.} PO, SU, PI, FV and SE  means ``Prominence'', ``Subitizing'', ``Prioritizing'', ``Free-viewing'', ``Searching'', respectively. 
    }
    \label{tab:supp_ablation_params}
    \vspace{-0.4cm}
\end{table}

\subsection{Discussion: Why do MLLMs work on HVS-related tasks?}
The design of MLLMs (particularly transformer-based architectures) allows approximations of human visual capabilities. 
{For Prominence, Subitizing and Prioritizing, \cite{tang2024cardiff} leverages MLLMs to derive visual saliency hierarchies as the guidance for saliency prediction, proving MLLMs' potential capacity to mimic human visual prioritization.}
{For Free-Viewing and Searching, \cite{yang2024unifying} can predict scanpaths by simulating gaze patterns using Transformer-based attention mechanisms.}
{For Prominence, Subitizing and Prioritizing, \cite{tian2024unsupervised} shows that human-like saliency can be simulated by attention layers inherently learning to weight salient regions of an input image, mirroring human prioritization of significant elements.}
{We highlight the value of benchmarking HVS alignment, as it benefits many tasks.}

\section{More Related Work}

\noindent\textbf{Human Visual System (HVS).}
The HVS has long been studied for its unique ability to process visual information efficiently and selectively. 
Computational modeling of HVS has gained significant traction in the fields of computer vision and cognitive neuroscience, aiming to replicate human-like attention and perception in artificial systems. 
\cite{itti1998model} shows how visual saliency guides human gaze patterns. Recent advancements in deep learning have incorporated human attention models into computer vision tasks, enabling better predictions of free-view human gaze~\cite{kummerer2016deepgaze,cornia2018predicting}. These approaches provide insight into how human cognition hierarchically processes visual information. HVS also demonstrates sequential and temporal fixation patterns, critical for understanding complex scenes~\cite{henderson2003human}. 
The study of the HVS has led to significant improvements and inspired new models in machine learning, such as attention models~\cite{vaswani2017attention}. It is crucial to conduct further research into the HVS due to its potential to advance the development of AGI. 

\noindent\textbf{Multimodal Large Language Models (MLLMs).}
MLLMs~\cite{alayrac2022flamingo} have emerged as a significant advancement in artificial intelligence, extending the capabilities of large language models to process and reason about both visual and textual information. 
By utilizing the open-source LLM~\cite{llama1,llama2,vicuna,glm} and the key idea of constructing visual instruction data, some powerful MLLMs have been proposed such as LLaVA~\cite{llava} and MiniGPT-4~\cite{minigpt4}. These models have shown their ability in general visual tasks.

Despite these advancements, questions about how MLLMs perceive and process visual information remain largely unexplored. It is unclear whether MLLMs fixate on regions of interest similar to humans or follow a comparable temporal sequence when perceiving images.  
Furthermore, further research in this area is hindered by the absence of standardized evaluation protocols and benchmarks.

\section{Application}
Content generation models better aligned with the HVS can produce more reasonable outputs. 
\ysqq{Take the prominent field for example, we design a Cropping-Based Prominence Enhancement to illustrate.}
Specifically, we examine how GPT-4o crops the image to enhance the prominence of one object: a photo. 
GPT-4o with a task-specific hint generates a reasonable analysis and successfully crops the image to highlight the photo, compared to the result without hint, demonstrating better alignment with HVS.
This can be directly applied to automated design, context-aware content generation, and visual storytelling.

\begin{figure}[t]
    \centering
    \includegraphics[width=0.9\linewidth]{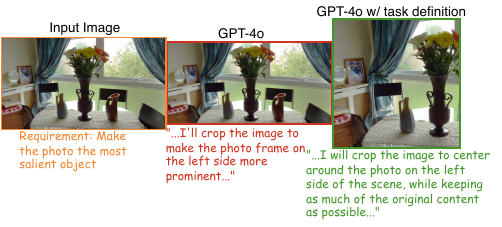}
    \vspace{-5mm}
    \caption{Application: Prominence Enhancement.}
    \label{fig:application_crop}
\end{figure}

{
    \small
    \bibliographystyle{ieeenat_fullname}
    \bibliography{main}

@String(CVPR= {IEEE Conf. Comput. Vis. Pattern Recog.})

@String(ICLR = {Int. Conf. Learn. Represent.})

@String(AAAI = {AAAI})

@String(CVPR  = {CVPR})

@String(ICLR  = {ICLR})

@article{itti1998model,
  title={A model of saliency-based visual attention for rapid scene analysis},
  author={Itti, Laurent and Koch, Christof and Niebur, Ernst},
  journal={IEEE Transactions on pattern analysis and machine intelligence},
  volume={20},
  number={11},
  pages={1254--1259},
  year={1998},
  publisher={Ieee}
}

@article{kummerer2016deepgaze,
  title={DeepGaze II: Reading fixations from deep features trained on object recognition},
  author={K{\"u}mmerer, Matthias and Wallis, Thomas SA and Bethge, Matthias},
  journal={arXiv preprint arXiv:1610.01563},
  year={2016}
}

@article{cornia2018predicting,
  title={Predicting human eye fixations via an lstm-based saliency attentive model},
  author={Cornia, Marcella and Baraldi, Lorenzo and Serra, Giuseppe and Cucchiara, Rita},
  journal={IEEE Transactions on Image Processing},
  volume={27},
  number={10},
  pages={5142--5154},
  year={2018},
  publisher={IEEE}
}

@article{henderson2003human,
  title={Human gaze control during real-world scene perception},
  author={Henderson, John M},
  journal={Trends in cognitive sciences},
  volume={7},
  number={11},
  pages={498--504},
  year={2003},
  publisher={Elsevier}
}

@article{alayrac2022flamingo,
  title={Flamingo: a visual language model for few-shot learning},
  author={Alayrac, Jean-Baptiste and Donahue, Jeff and Luc, Pauline and Miech, Antoine and Barr, Iain and Hasson, Yana and Lenc, Karel and Mensch, Arthur and Millican, Katherine and Reynolds, Malcolm and others},
  journal={Advances in neural information processing systems},
  volume={35},
  pages={23716--23736},
  year={2022}
}

@article{desimone1995neural,
  title={Neural mechanisms of selective visual attention},
  author={Desimone, Robert and Duncan, John and others},
  journal={Annual review of neuroscience},
  volume={18},
  number={1},
  pages={193--222},
  year={1995}
}

@inproceedings{vaswani2017attention,
 author = {Vaswani, Ashish and Shazeer, Noam and Parmar, Niki and Uszkoreit, Jakob and Jones, Llion and Gomez, Aidan N and Kaiser, \L ukasz and Polosukhin, Illia},
 booktitle = {Advances in Neural Information Processing Systems},
 editor = {I. Guyon and U. Von Luxburg and S. Bengio and H. Wallach and R. Fergus and S. Vishwanathan and R. Garnett},
 pages = {},
 publisher = {Curran Associates, Inc.},
 title = {Attention is All you Need},
 volume = {30},
 year = {2017}
}

@article{liu2023mmbench,
  title={MMBench: Is your multi-modal model an all-around player?},
  author={Liu, Yuan and Duan, Haodong and Zhang, Yuanhan and others},
  journal={arXiv preprint arXiv:2307.06281},
  year={2023}
}

@article{li2023seedbench,
  title={SEEDBench: Benchmarking Multimodal LLMs with Generative Comprehension},
  author={Li, Bohao and Wang, Rui and Ge, Yuying and others},
  journal={arXiv preprint arXiv:2307.16125},
  year={2023}
}

@article{duan2024vlmevalkit,
  title={VLMEvalKit: An Open-Source Toolkit for Evaluating Large Multi-Modality Models},
  author={Duan, Haodong and Yang, Junming and Qiao, Yuxuan and others},
  journal={arXiv preprint arXiv:2407.11691},
  year={2024}
}

@inproceedings{singh2019towards,
  title={Towards VQA models that can read},
  author={Singh, Amanpreet and others},
  booktitle={CVPR},
  year={2019}
}

@inproceedings{mathew2021docvqa,
  title={DocVQA: A dataset for VQA on document images},
  author={Mathew, Minesh and others},
  booktitle={WACV},
  year={2021}
}

@article{chen2024far_internvl2,
  title={How far are we to gpt-4v? closing the gap to commercial multimodal models with open-source suites},
  author={Chen, Zhe and Wang, Weiyun and Tian, Hao and Ye, Shenglong and Gao, Zhangwei and Cui, Erfei and Tong, Wenwen and Hu, Kongzhi and Luo, Jiapeng and Ma, Zheng and others},
  journal={arXiv preprint arXiv:2404.16821},
  year={2024}
}

@article{achiam2023gpt4o,
  title={Gpt-4 technical report},
  author={Achiam, Josh and Adler, Steven and Agarwal, Sandhini and Ahmad, Lama and Akkaya, Ilge and Aleman, Florencia Leoni and Almeida, Diogo and Altenschmidt, Janko and Altman, Sam and Anadkat, Shyamal and others},
  journal={arXiv preprint arXiv:2303.08774},
  year={2023}
}

@article{minigpt4,
  title={MiniGPT-4: Enhancing Vision-Language Understanding with Advanced Large Language Models},
  author={Deyao Zhu and Jun Chen and Xiaoqian Shen and Xiang Li and Mohamed Elhoseiny},
  journal={ArXiv},
  year={2023},
  volume={abs/2304.10592},
}

@article{ye2024mplug,
  title={mplug-owl3: Towards long image-sequence understanding in multi-modal large language models},
  author={Ye, Jiabo and Xu, Haiyang and Liu, Haowei and Hu, Anwen and Yan, Ming and Qian, Qi and Zhang, Ji and Huang, Fei and Zhou, Jingren},
  journal={arXiv preprint arXiv:2408.04840},
  year={2024}
}

@article{wang2024qwen2vl,
  title={Qwen2-vl: Enhancing vision-language model's perception of the world at any resolution},
  author={Wang, Peng and Bai, Shuai and Tan, Sinan and Wang, Shijie and Fan, Zhihao and Bai, Jinze and Chen, Keqin and Liu, Xuejing and Wang, Jialin and Ge, Wenbin and others},
  journal={arXiv preprint arXiv:2409.12191},
  year={2024}
}

@inproceedings{llava,
  title={Visual Instruction Tuning},
  author={Haotian Liu and Chunyuan Li and Qingyang Wu and Yong Jae Lee},
  booktitle={NeurIPS},
  year={2023}
}

@article{llama1,
  title={LLaMA: Open and Efficient Foundation Language Models},
  author={Hugo Touvron and Thibaut Lavril and Gautier Izacard and Xavier Martinet and Marie-Anne Lachaux and Timoth{\'e}e Lacroix and Baptiste Rozi{\`e}re and Naman Goyal and Eric Hambro and Faisal Azhar and Aurelien Rodriguez and Armand Joulin and Edouard Grave and Guillaume Lample},
  journal={ArXiv},
  year={2023},
  volume={abs/2302.13971},
}

@article{llama2,
  title={Llama 2: Open Foundation and Fine-Tuned Chat Models},
  author={Hugo Touvron and Louis Martin and Kevin R. Stone and Peter Albert and Amjad Almahairi and Yasmine Babaei and Nikolay Bashlykov and Soumya Batra and Prajjwal Bhargava and Shruti Bhosale and Daniel M. Bikel and Lukas Blecher and Cristian Cant{\'o}n Ferrer and Moya Chen and Guillem Cucurull and David Esiobu and Jude Fernandes and Jeremy Fu and Wenyin Fu and Brian Fuller and Cynthia Gao and Vedanuj Goswami and Naman Goyal and Anthony S. Hartshorn and Saghar Hosseini and Rui Hou and Hakan Inan and Marcin Kardas and Viktor Kerkez and Madian Khabsa and Isabel M. Kloumann and A. V. Korenev and Punit Singh Koura and Marie-Anne Lachaux and Thibaut Lavril and Jenya Lee and Diana Liskovich and Yinghai Lu and Yuning Mao and Xavier Martinet and Todor Mihaylov and Pushkar Mishra and Igor Molybog and Yixin Nie and Andrew Poulton and Jeremy Reizenstein and Rashi Rungta and Kalyan Saladi and Alan Schelten and Ruan Silva and Eric Michael Smith and R. Subramanian and Xia Tan and Binh Tang and Ross Taylor and Adina Williams and Jian Xiang Kuan and Puxin Xu and Zhengxu Yan and Iliyan Zarov and Yuchen Zhang and Angela Fan and Melanie Kambadur and Sharan Narang and Aurelien Rodriguez and Robert Stojnic and Sergey Edunov and Thomas Scialom},
  journal={ArXiv},
  year={2023},
  volume={abs/2307.09288},
}

@misc{vicuna,
    title={Vicuna: An open-source chatbot impressing gpt-4 with 90\% chatgpt quality},
    author={Vicuna Team},
    howpublished = {\url{https://vicuna.lmsys. org/}},
    year={2023}
}

@inproceedings{glm,
  title={GLM-130B: An Open Bilingual Pre-trained Model},
  author={Aohan Zeng and Xiao Liu and Zhengxiao Du and Zihan Wang and Hanyu Lai and Ming Ding and Zhuoyi Yang and Yifan Xu and Wendi Zheng and Xiao Xia and Weng Lam Tam and Zixuan Ma and Yufei Xue and Jidong Zhai and Wenguang Chen and P. Zhang and Yuxiao Dong and Jie Tang},
  booktitle={ICLR},
  year={2022},
}

@inproceedings{lu2024mathvista,
  author    = {Lu, Pan and Bansal, Hritik and Xia, Tony and Liu, Jiacheng and Li, Chunyuan and Hajishirzi, Hannaneh and Cheng, Hao and Chang, Kai-Wei and Galley, Michel and Gao, Jianfeng},
  title     = {MathVista: Evaluating Mathematical Reasoning of Foundation Models in Visual Contexts},
  booktitle={International Conference on Learning Representations (ICLR)},
  year      = {2024}
}

@inproceedings{deng2024advancing,
  title={Advancing Saliency Ranking with Human Fixations: Dataset Models and Benchmarks},
  author={Deng, Bowen and Song, Siyang and French, Andrew P and Schluppeck, Denis and Pound, Michael P},
  booktitle={Proceedings of the IEEE/CVF Conference on Computer Vision and Pattern Recognition},
  pages={28348--28357},
  year={2024}
}

@article{pei2022oqtr,
  title={Transformer-based Efficient Salient Instance Segmentation Networks with Orientative Query},
  author={Pei, Jialun and Cheng, Tianyang and Tang, He and Chen, Chuanbo},
  journal={IEEE Transactions on Multimedia},
  year={2022},
  publisher={IEEE}
}

@inproceedings{chen2022characterizing, 
title={Characterizing Target-Absent Human Attention}, 
author={Chen, Yupei and Yang, Zhibo and Chakraborty, Souradeep and Mondal, Sounak and Ahn, Seoyoung and Samaras, Dimitris and Hoai, Minh and Zelinsky, Gregory}, 
booktitle={Proceedings of the IEEE/CVF Conference on Computer Vision and Pattern Recognition Workshops}, 
pages={5031--5040}, 
year={2022} 
}

@InProceedings{Yang_2020_CVPR,
author = {Yang, Zhibo and Huang, Lihan and Chen, Yupei and Wei, Zijun and Ahn, Seoyoung and Zelinsky, Gregory and Samaras, Dimitris and Hoai, Minh},
title = {Predicting Goal-Directed Human Attention Using Inverse Reinforcement Learning},
booktitle = {The IEEE/CVF Conference on Computer Vision and Pattern Recognition (CVPR)},
month = {June},
year = {2020}
}

@inproceedings{zhang2015salient,
  title={Salient object subitizing},
  author={Zhang, Jianming and Ma, Shugao and Sameki, Mehrnoosh and Sclaroff, Stan and Betke, Margrit and Lin, Zhe and Shen, Xiaohui and Price, Brian and Mech, Radomir},
  booktitle={Proceedings of the IEEE Conference on Computer Vision and Pattern Recognition},
  pages={4045--4054},
  year={2015}
}

@InProceedings{yang2024unify,
  author = {Yang, Zhibo and Mondal, Sounak and Ahn, Seoyoung and Xue, Ruoyu and Zelinsky, Gregory and Hoai, Minh and Samaras, Dimitris},
  title = {Unifying Top-down and Bottom-up Scanpath Prediction Using Transformers},
  booktitle = {The IEEE Conference on Computer Vision and Pattern Recognition (CVPR)},
  month = {June},
  year = {2024}
}

@misc{lu2024deepseekvl,
      title={DeepSeek-VL: Towards Real-World Vision-Language Understanding},
      author={Haoyu Lu and Wen Liu and Bo Zhang and Bingxuan Wang and Kai Dong and Bo Liu and Jingxiang Sun and Tongzheng Ren and Zhuoshu Li and Hao Yang and Yaofeng Sun and Chengqi Deng and Hanwei Xu and Zhenda Xie and Chong Ruan},
      year={2024},
      eprint={2403.05525},
      archivePrefix={arXiv},
      primaryClass={cs.AI}
}

@article{li2024llava_onevision,
  title={Llava-onevision: Easy visual task transfer},
  author={Li, Bo and Zhang, Yuanhan and Guo, Dong and Zhang, Renrui and Li, Feng and Zhang, Hao and Zhang, Kaichen and Li, Yanwei and Liu, Ziwei and Li, Chunyuan},
  journal={arXiv preprint arXiv:2408.03326},
  year={2024}
}

@misc{laurencon2023obelics_idefics,
      title={OBELICS: An Open Web-Scale Filtered Dataset of Interleaved Image-Text Documents},
      author={Hugo Laurençon and Lucile Saulnier and Léo Tronchon and Stas Bekman and Amanpreet Singh and Anton Lozhkov and Thomas Wang and Siddharth Karamcheti and Alexander M. Rush and Douwe Kiela and Matthieu Cord and Victor Sanh},
      year={2023},
      eprint={2306.16527},
      archivePrefix={arXiv},
      primaryClass={cs.IR}
}

@inproceedings{jarodzka2010vector,
  title={A vector-based, multidimensional scanpath similarity measure},
  author={Jarodzka, Halszka and Holmqvist, Kenneth and Nystr{\"o}m, Marcus},
  booktitle={Proceedings of the 2010 symposium on eye-tracking research \& applications},
  pages={211--218},
  year={2010}
}

@article{yao2024minicpm,
  title={MiniCPM-V: A GPT-4V Level MLLM on Your Phone},
  author={Yao, Yuan and Yu, Tianyu and Zhang, Ao and Wang, Chongyi and Cui, Junbo and Zhu, Hongji and Cai, Tianchi and Li, Haoyu and Zhao, Weilin and He, Zhihui and others},
  journal={arXiv preprint arXiv:2408.01800},
  year={2024}
}

@article{team2024gemini,
  title={Gemini 1.5: Unlocking multimodal understanding across millions of tokens of context},
  author={Team, Gemini and Georgiev, Petko and Lei, Ving Ian and Burnell, Ryan and Bai, Libin and Gulati, Anmol and Tanzer, Garrett and Vincent, Damien and Pan, Zhufeng and Wang, Shibo and others},
  journal={arXiv preprint arXiv:2403.05530},
  year={2024}
}

@article{coco_caption,
  title={Microsoft coco captions: Data collection and evaluation server},
  author={Chen, Xinlei and Fang, Hao and Lin, Tsung-Yi and Vedantam, Ramakrishna and Gupta, Saurabh and Doll{\'a}r, Piotr and Zitnick, C Lawrence},
  journal={arXiv:1504.00325},
  year={2015}
}

@inproceedings{hudson2019gqa,
  title={Gqa: A new dataset for real-world visual reasoning and compositional question answering},
  author={Hudson, Drew A and Manning, Christopher D},
  booktitle={Proceedings of the IEEE/CVF conference on computer vision and pattern recognition},
  pages={6700--6709},
  year={2019}
}

@article{laurenccon2024building,
  title={Building and better understanding vision-language models: insights and future directions},
  author={Lauren{\c{c}}on, Hugo and Marafioti, Andr{\'e}s and Sanh, Victor and Tronchon, L{\'e}o},
  journal={arXiv preprint arXiv:2408.12637},
  year={2024}
}

@article{li2023evaluating,
  title={Evaluating object hallucination in large vision-language models},
  author={Li, Yifan and Du, Yifan and Zhou, Kun and Wang, Jinpeng and Zhao, Wayne Xin and Wen, Ji-Rong},
  journal={arXiv preprint arXiv:2305.10355},
  year={2023}
}

@article{zhou2021rgb,
  title={RGB-D salient object detection: A survey},
  author={Zhou, Tao and Fan, Deng-Ping and Cheng, Ming-Ming and Shen, Jianbing and Shao, Ling},
  journal={Computational Visual Media},
  volume={7},
  pages={37--69},
  year={2021},
  publisher={Springer}
}

@article{borji2014salient,
  title={What is a salient object? A dataset and a baseline model for salient object detection},
  author={Borji, Ali},
  journal={IEEE Transactions on Image Processing},
  volume={24},
  number={2},
  pages={742--756},
  year={2014},
  publisher={IEEE}
}

@inproceedings{cheng2014depth,
  title={Depth enhanced saliency detection method},
  author={Cheng, Yupeng and Fu, Huazhu and Wei, Xingxing and Xiao, Jiangjian and Cao, Xiaochun},
  booktitle={Proceedings of international conference on internet multimedia computing and service},
  pages={23--27},
  year={2014}
}

@inproceedings{cerf2008using,
  title={Using semantic content as cues for better scanpath prediction},
  author={Cerf, Moran and Frady, E Paxon and Koch, Christof},
  booktitle={Proceedings of the 2008 symposium on Eye tracking research \& applications},
  pages={143--146},
  year={2008}
}

@article{jiang2016learning,
  title={Learning to predict sequences of human visual fixations},
  author={Jiang, Ming and Boix, Xavier and Roig, Gemma and Xu, Juan and Van Gool, Luc and Zhao, Qi},
  journal={IEEE transactions on neural networks and learning systems},
  volume={27},
  number={6},
  pages={1241--1252},
  year={2016},
  publisher={IEEE}
}

@inproceedings{siris2020inferring,
  title={Inferring attention shift ranks of objects for image saliency},
  author={Siris, Avishek and Jiao, Jianbo and Tam, Gary KL and Xie, Xianghua and Lau, Rynson WH},
  booktitle={Proceedings of the IEEE/CVF conference on computer vision and pattern recognition},
  pages={12133--12143},
  year={2020}
}

@article{koehler2014saliency,
  title={What do saliency models predict?},
  author={Koehler, Kathryn and Guo, Fei and Zhang, Sheng and Eckstein, Miguel P},
  journal={Journal of vision},
  volume={14},
  number={3},
  pages={14--14},
  year={2014},
  publisher={The Association for Research in Vision and Ophthalmology}
}

@InProceedings{wu2024v,
    author    = {Wu, Penghao and Xie, Saining},
    title     = {V?: Guided Visual Search as a Core Mechanism in Multimodal LLMs},
    booktitle = {Proceedings of the IEEE/CVF Conference on Computer Vision and Pattern Recognition (CVPR)},
    month     = {June},
    year      = {2024},
    pages     = {13084-13094}
}

@incollection{henderson2013scene,
  title={Scene perception for psycholinguists},
  author={Henderson, John M and Ferreira, Fernanda},
  booktitle={The interface of language, vision, and action},
  pages={1--58},
  year={2013},
  publisher={Psychology Press}
}

@article{trick1994small,
  title={Why are small and large numbers enumerated differently? A limited-capacity preattentive stage in vision.},
  author={Trick, Lana M and Pylyshyn, Zenon W},
  journal={Psychological review},
  volume={101},
  number={1},
  pages={80},
  year={1994},
  publisher={American Psychological Association}
}

@article{eckstein2011visual,
  title={Visual search: A retrospective},
  author={Eckstein, Miguel P},
  journal={Journal of vision},
  volume={11},
  number={5},
  pages={14--14},
  year={2011},
  publisher={The Association for Research in Vision and Ophthalmology}
}

@inproceedings{yang2024unifying,
  title={Unifying Top-down and Bottom-up Scanpath Prediction Using Transformers},
  author={Yang, Zhibo and Mondal, Sounak and Ahn, Seoyoung and Xue, Ruoyu and Zelinsky, Gregory and Hoai, Minh and Samaras, Dimitris},
  booktitle={Proceedings of the IEEE/CVF Conference on Computer Vision and Pattern Recognition},
  pages={1683--1693},
  year={2024}
}

@inproceedings{tian2024unsupervised,
  title={Unsupervised Salient Instance Detection},
  author={Tian, Xin and Xu, Ke and Lau, Rynson},
  booktitle={Proceedings of the IEEE/CVF Conference on Computer Vision and Pattern Recognition},
  pages={2702--2712},
  year={2024}
}

@article{tang2024cardiff,
  title = {CaRDiff: Video Salient Object Ranking Chain of Thought Reasoning for Saliency Prediction with Diffusion, AAAI2025},
  author = {Tang, Yunlong and Zhan, Gen and Yang, Li and Liao, Yiting and Xu, Chenliang},
  journal = {AAAI Conference on Artificial Intelligence (AAAI)},
  year = {2025},
}

@inproceedings{li2014secrets,
  title={The secrets of salient object segmentation},
  author={Li, Yin and Hou, Xiaodi and Koch, Christof and Rehg, James M and Yuille, Alan L},
  booktitle={Proceedings of the IEEE conference on computer vision and pattern recognition},
  pages={280--287},
  year={2014}
}

@article{beck2009top,
  title={Top-down and bottom-up mechanisms in biasing competition in the human brain},
  author={Beck, Diane M and Kastner, Sabine},
  journal={Vision research},
  volume={49},
  number={10},
  pages={1154--1165},
  year={2009},
  publisher={Elsevier}
}

@misc{ernie2025technicalreport,
      title={ERNIE 4.5 Technical Report},
      author={Baidu ERNIE Team},
      year={2025},
      eprint={},
      archivePrefix={arXiv},
      primaryClass={cs.CL},
      url={}
}

@misc{vteam2025glm45vglm41vthinkingversatilemultimodal,
      title={GLM-4.5V and GLM-4.1V-Thinking: Towards Versatile Multimodal Reasoning with Scalable Reinforcement Learning}, 
      author={V Team and Wenyi Hong and Wenmeng Yu and Xiaotao Gu and Guo Wang and Guobing Gan and Haomiao Tang and Jiale Cheng and Ji Qi and Junhui Ji and Lihang Pan and Shuaiqi Duan and Weihan Wang and Yan Wang and Yean Cheng and Zehai He and Zhe Su and Zhen Yang and Ziyang Pan and Aohan Zeng and Baoxu Wang and Bin Chen and Boyan Shi and Changyu Pang and Chenhui Zhang and Da Yin and Fan Yang and Guoqing Chen and Jiazheng Xu and Jiale Zhu and Jiali Chen and Jing Chen and Jinhao Chen and Jinghao Lin and Jinjiang Wang and Junjie Chen and Leqi Lei and Letian Gong and Leyi Pan and Mingdao Liu and Mingde Xu and Mingzhi Zhang and Qinkai Zheng and Sheng Yang and Shi Zhong and Shiyu Huang and Shuyuan Zhao and Siyan Xue and Shangqin Tu and Shengbiao Meng and Tianshu Zhang and Tianwei Luo and Tianxiang Hao and Tianyu Tong and Wenkai Li and Wei Jia and Xiao Liu and Xiaohan Zhang and Xin Lyu and Xinyue Fan and Xuancheng Huang and Yanling Wang and Yadong Xue and Yanfeng Wang and Yanzi Wang and Yifan An and Yifan Du and Yiming Shi and Yiheng Huang and Yilin Niu and Yuan Wang and Yuanchang Yue and Yuchen Li and Yutao Zhang and Yuting Wang and Yu Wang and Yuxuan Zhang and Zhao Xue and Zhenyu Hou and Zhengxiao Du and Zihan Wang and Peng Zhang and Debing Liu and Bin Xu and Juanzi Li and Minlie Huang and Yuxiao Dong and Jie Tang},
      year={2025},
      eprint={2507.01006},
      archivePrefix={arXiv},
      primaryClass={cs.CV},
      url={https://arxiv.org/abs/2507.01006}, 
}
}

\end{document}